\documentclass[journal,comsoc]{IEEEtran}

\usepackage[T1]{fontenc}
\ifCLASSINFOpdf
\else 
\fi
\usepackage{amsmath}
\usepackage[cmintegrals]{newtxmath}
\usepackage{boldline,multirow}
\usepackage{graphicx}
\usepackage{subfigure}
\usepackage[noadjust]{cite}
\usepackage{enumerate}
\usepackage{diagbox}
\usepackage{makecell}
\usepackage{blindtext}
\usepackage{enumitem}
\usepackage{multirow}
\usepackage[hang,flushmargin]{footmisc}

\usepackage[linesnumbered,ruled,vlined]{algorithm2e}

\SetCommentSty{mycommfont}
\makeatletter
\newcommand{\thickhline}{%
	\noalign {\ifnum 0=`}\fi \hrule height 1pt
	\futurelet \reserved@a \@xhline
}
\newcolumntype{"}{@{\hskip\tabcolsep\vrule width 1pt\hskip\tabcolsep}}
\makeatother

\usepackage[dvipsnames,table,xcdraw]{xcolor}

\let\oldnl\nl
\newcommand{\nonl}{\renewcommand{\nl}{\let\nl\oldnl}}

\usepackage{lipsum}
\usepackage{fancyhdr}

\begin{document}
	
	\title{Evolutionary Ensemble Learning for Multivariate Time Series Prediction
		\thanks{This work is supported by Buildings Engineered for Sustainability research project, funded by RMIT Sustainable Urban Precincts Program.}}
	
	\author{Hui~Song,
		A. K. ~Qin,
		Flora D. Salim
		\thanks{Hui Song, Flora D. Salim, Computer Science and Information Technology, School of Science, RMIT University, Melbourne, VIC 3000 (hui.song@rmit.edu.au, flora.salim@rmit.edu.au).}
		\thanks{A. K. Qin, Department of Computer Science and Software Engineering, Swinburne University of Technology, Hawthorn, VIC 3122 (kqin@swin.edu.au).}}
	
	\markboth{IEEE Transactions on Systems, Man, and Cybernetics: Systems}%
	{Shell \MakeLowercase{\textit{et al.}}: Bare Demo of IEEEtran.cls for IEEE Communications Society Journals}
	
	\maketitle
	
	\begin{abstract}
		Multivariate time series (MTS) prediction plays a key role in many fields such as finance, energy and transport, where each individual time series corresponds to the data collected from a certain data source, so-called channel. A typical pipeline of building an MTS prediction model (PM) consists of selecting a subset of channels among all available ones, extracting features from the selected channels, and building a PM based on the extracted features, where each component involves certain optimization tasks, i.e., selection of channels, feature extraction (FE) methods, and PMs as well as configuration of the selected FE method and PM. Accordingly, pursuing the best prediction performance corresponds to optimizing the pipeline by solving all of its involved optimization problems. This is a non-trivial task due to the vastness of the solution space. Different from most of the existing works which target at optimizing certain components of the pipeline, we propose a novel evolutionary ensemble learning framework to optimize the entire pipeline in a holistic manner. In this framework, a specific pipeline is encoded as a candidate solution and a multi-objective evolutionary algorithm is applied under different population sizes to produce multiple Pareto optimal sets (POSs). Finally, selective ensemble learning is designed to choose the optimal subset of solutions from the POSs and combine them to yield final prediction by using greedy sequential selection and least square methods. We implement the proposed framework and evaluate our implementation on two real-world applications, i.e., electricity consumption prediction and air quality prediction. The performance comparison with state-of-the-art techniques demonstrates the superiority of the proposed approach.
		
	\end{abstract}
	
	\begin{IEEEkeywords}
		Multivariate time series, multi-objective evolutionary algorithm, selective ensemble learning, least square.
		
	\end{IEEEkeywords}
	
	\IEEEpeerreviewmaketitle
	
	\section{Introduction}
	Nowadays, the data generated in many sectors, e.g., transportation, finance, energy, and health \cite{de200625}, are usually in the form of multiple time series (MTS), where each individual time series (TS) typically corresponds to the data collected from a certain data source, so-called channel. Over the years, many approaches have been proposed to solve various MTS prediction tasks \cite{gangopadhyay2020spatiotemporal, hu2020multistage, qin2017dual, du2020multivariate}. A typical pipeline for MTS prediction is composed of channel selection (CS), feature extraction (FE) and predictive modelling, where certain appropriate methods and their associated parameter settings need to be manually specified for each component in the pipeline. To achieve performance optimality, it is desirable to automatically search the most effective methods armed with the best calibrated parameter values for any given prediction task. This corresponds to solving a complex optimization problem. In this problem, the sub-problem of CS has a large discrete search space that grows exponentially as the number of channels increases. For applying FE to the selected channels, there exist a huge number of candidate FE techniques such as \cite{luo2015piecewise,fan2014development, fu2011review, QIN2005613} with each having some associated parameters to be set, e.g., the time window (TW) size \cite{song2016multivariate}. Choosing among them the most effective technique with its parameters best calibrated corresponds to dealing with a challenging optimization sub-problem that contains both discrete (for techniques) and continuous (for parameters) decision variables. Further, there is another challenging optimization sub-problem, with decision variables of mixed types, of finding the best prediction model (PM) among numerous existing candidates such as \cite{de200625, QIN2005773} while calibrating its parameters to the best to fit into the pipeline for the purpose of achieving the best prediction performance.
	
	Solving this optimization problem poses a great challenge due to its non-differentiable and mixed-integer nature which disables the feasibility of employing gradient descent methods. Evolutionary algorithms (EAs) \cite{DEJONG2016}, as a well-known family of nature-inspired population-based stochastic search techniques with strong competency in dealing with non-convex and non-differential optimization problems \cite{GONG2016158}, provide a native solution to address this problem. They have been successfully used to address some parts of this problem, e.g., optimization of the PM \cite{floreano2008neuroevolution, stanley2019designing}. However, to the best of our knowledge, none of the existing approaches has addressed this problem as a whole, i.e., simultaneously solving all three sub-problems in terms of CS, FE and prediction in a holistic manner.
	
	Due to high nonlinearity and complexity, this optimization problem may have numerous local optima. Although EAs are competent to find superior local optima (and even the global one), the generalization performance may not correspond to the optimality achieved via the training process, e.g., overfitting may occur. Further, different local optima in the solution space, e.g., those with different selected channels, FE methods or PMs, may have the similar quality in terms of solving the optimization problem but result in the distinct generalization performance on the test set. To address this issue, evolutionary ensemble learning (EEL) \cite{chen2010multiobjective, chandra2006ensemble} provides a native solution. It typically employs multi-objective EAs (MOEAs) \cite{zhang2007moea} to produce a set of non-dominated solutions, so-called the Pareto optimal set (POS), by considering the prediction performance and diversity of solutions as two conflicting objectives, and combines some or all of the solutions in the Pareto optimal set in a certain way to yield the final prediction result. It has been widely demonstrated that EEL may lead to improved generalisation \cite{zhang2017multiobjective, song2018evolutionary}. However, this kind of approaches may be highly sensitive to some algorithmic parameters like the population size. Specifically, using different population sizes may lead to distinct POSs. Further, the gap between the estimated prediction performance (employed as one objective in MOEAs) and the generalisation performance and the way to create the ensemble model from the Pareto optimal set may also influence the ultimate performance.
	
	To address the above issues, we propose a novel EEL based on  selective ensemble learning (SEL) framework for MTS prediction. In this framework, the solution space covers all possible configurations of the whole MTS prediction pipeline, where any specific setting regarding selection of channels, selection and configuration of FE methods and selection and configuration of PMs corresponds to a candidate solution in the solution space. The MOEAs under multiple different population sizes are run independently, subjected to the prediction performance and diversity of solutions as two objectives, to search diverse optimal pipeline configurations, producing multiple POSs corresponding to different populating sizes. SEL is designed based on the generated POSs from which some solutions are selected in a task-oriented manner and used to build the final ensemble PM. Notably, the proposed EEL framework has three key features which differentiate it from the existing EEL frameworks, i.e., holistic solution representation, multiple POSs, and creation of a selective ensemble PM in a task-oriented way.
	
	We implement the EEL framework by employing MOEA/D  \cite{zhang2007moea} as the MOEA, where we use five-fold cross-validation accuracy, which can better estimate generalisation, as an objective regarding prediction performance. As for solution representation, each evolved Pareto front (PF) is composed of a set of non-dominated optimal solutions (models). Each of them encodes the selection of channels, FE methods and TWs, and PMs as well as configuration of the selected FE methods and PM. Then, a wrapper-based feature selection method \cite{chandrashekar2014survey} and the least square (LS) regressor \cite{li2016ensemble} is applied to select a subset of the solutions from multiple POSs obtained by MOEA/D under different population sizes for the ensemble modelling. Accordingly, a LS-based selective ensemble PM is established, where the outputs of all the selected solutions are used as the input for the LS regressor to produce the final prediction result. We evaluate the performance of our implementation on two real-world MTS prediction tasks by conducting the ablation study of the key components in the proposed EEL and comparing it with several state-of-the-art MTS prediction techniques. 
	
	The rest of this paper is organized as follows. Section~\ref{Background} illustrates the background of related techniques. The related work is depicted in Section~\ref{Related_work}. Section~\ref{Proposed_method} introduces the proposed framework and its implementation. Experimental results are reported and discussed in Section~\ref{Experiments}. Section~\ref{Conclusions} concludes the paper with some future work being mentioned.
	
	\section{Background}\label{Background}
	
	\subsection{Multi-objective Optimization}
	
	For many real-world applications, there are always more than one conflicting objectives to be optimized simultaneously. Problems like this are called multi-objective optimization problems (MOPs), defined as follows:
	\begin{align}
		\label{MOP}
		\left\{\begin{array}{rl}
			\text{Minimize:} & \textbf{F}(\text{x})=(f_1(\text{x}), f_2(\text{x}),..., f_m(\text{x}))^T \\ 
			\text{Subject to:} & x\in \Omega  
		\end{array}\right.
	\end{align}
	where $\Omega \subset \mathbb{R}^n$ represents the decision space and $\text{x}=(x_1, x_2,..., x_n) \in \Omega$ is a decision variable with respect to a solution for a specific MOP. $\textbf{F}(\text{x}): \Omega \to \mathbb{R}^m$ denotes the $m$-dimensional objective vector of the solution $\text{x}$.
	
	For a specific MOP, assuming $\text{x}_A$ and $\text{x}_B$ are its two solutions, if and only if $f_i(\text{x}_A) \leq f_i(\text{x}_B), \forall i\in \{1,..., m\}$ and there is a $j\in \{1,..., m\}$ satisfying $f_j(\text{x}_A) < f_j(\text{x}_B)$, $\text{x}_A$ dominates $\text{x}_B$ (i.e., $\text{x}_A\prec \text{x}_B$). A solution $\text{x}^* \in \Omega$ is called Pareto optimum if there is no other solution that dominates $\text{x}^*$. The set of all Pareto optimal solutions is called Pareto optimal set (PS) and Pareto optimal front is defined as the corresponding objective vectors of the solutions in the PF. Therefore, to solve a MOP is to find its PS. 
	
	\subsection{Extreme Learning Machine}
	
\begin{figure}[h!]\centering
		\subfigure[ELM]
		{\centering\scalebox{0.65}[0.65]
			{\includegraphics{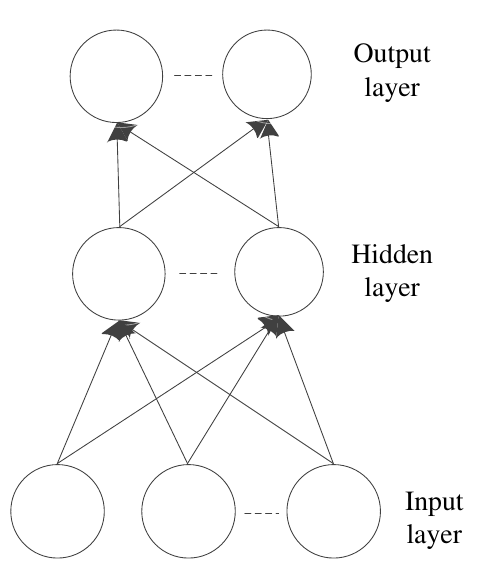}}\label{ELM_structure}}
		\subfigure[RVFL]
		{\centering\scalebox{0.65}[0.65]
			{\includegraphics{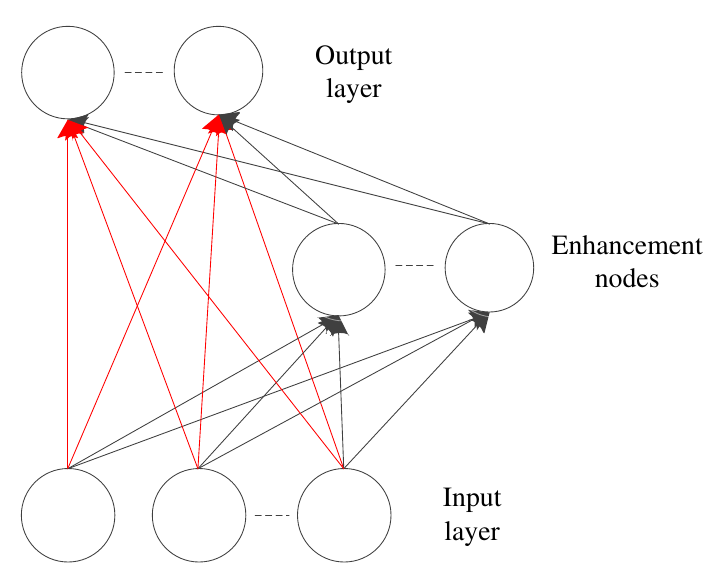}}\label{RVFL_structure}}
		\caption{The structures of ELM and RVFL.} \label{ELM_RVFL}
	\end{figure}

	Extreme learning machine (ELM), proposed by Huang \textit{et al}. \cite{huang2006extreme}, mainly focuses on solving the drawbacks caused by gradient descent based algorithms. ELM is based on single hidden layer feedforward neural network (SLFN) architecture and includes three different layers: input layer, hidden layer and output layer, shown as Fig.~\ref{ELM_structure}. The hidden bias and weights for connecting the input layer and hidden layer are generated randomly and maintained through the whole training process.
	
	Assuming dataset $(\textbf{x}_i,\textbf{y}_i)$ with a set of $M$ distinct samples, satisfied $\textbf{x}_i\in{\mathcal{R}^{d1}}$ and $\textbf{y}_i\in{\mathcal{R}^{d2}}$, a SLFN with $N$ hidden neurons can be mathematically formulated as:
	
	\begin{equation}
		\label{ELM1}
		\sum\limits_{i=1}^{N}{\boldsymbol{\beta}}_if(\textbf{w}_i^T\textbf{x}_j+b_i), 1\le{j}\le{M}\vspace{-0.1cm}
	\end{equation}
	where $f$ is the activation function; $\textbf{w}_i$ represents the weights for connecting input layer and hidden layer; $b_i$ is bias and $\mathbf{\beta}_i$ is the output weight.
	
	In ELM, the structure perfectly approximates to the given output data:
	\begin{equation}
		\label{ELM2}
		\sum\limits_{i=1}^{N}{\boldsymbol{\beta}}_if(\textbf{w}_i^T\textbf{x}_j+b_i)=\textbf{y}_j, 1\le{j}\le{M}\vspace{-0.1cm}
	\end{equation}	
	which can be written as $\textbf{HB}=\textbf{Y}$, the matrix $\textbf{H}$ can be represented as:
	\begin{equation}
		\label{ELM3}
		\textbf{H} = 
		\begin{pmatrix}
			f(\textbf{w}_1^T\textbf{x}_1+b_1) & \cdots & f(\textbf{w}_N^T\textbf{x}_1+b_N) \\
			\cdots  & \cdots & \cdots  \\
			f(\textbf{w}_1^T\textbf{x}_M+b_1) & \cdots & f(\textbf{w}_N^T\textbf{x}_M+b_N) 
		\end{pmatrix}
	\end{equation}
	
	$\textbf{B}=(\boldsymbol{\beta}_1^T, \boldsymbol{\beta}_2^T,...,\boldsymbol{\beta}_N^T)^T$  and $\textbf{Y}=(y_1^T,y_2^T,...,y_M^T)^T$.
	
	The output weight $\textbf{B}$ is calculated by $\textbf{B}=\textbf{H}^+\textbf{Y}$, and $\textbf{H}^+$ is a Moore-Penrose generalized inverse of $\textbf{H}$ \cite{rao1971generalized}. Theoretical proofs and a more thorough presentation of the ELM algorithm are detailed in the original paper \cite{huang2006extreme}.
	
	\subsection{Random Vector Functional Link Neural Network} Random vector functional link (RVFL) neural network was proposed by Pao and Takefuji \cite{pao1992functional}. As illustrated in Fig.~\ref{RVFL_structure}, the weights $\textbf{w}_i$ from the input to the enhancement nodes are randomly generated like ELM, therefore the output of the $i$th enhancement node (the number of nodes is $N$) is $f(\textbf{w}_i^T\textbf{x}_j+b_i), 1\le{j}\le{N}, 1\le{j}\le{M}$. Similar with ELM, the structure can approximate to the given output data:
	\begin{equation}
		\label{RVFL1}
		\sum\limits_{i=1}^{N}{\boldsymbol{\beta}}_if(\textbf{w}_i^T\textbf{x}_j+b_i)+\sum\limits_{i=N+1}^{N+d_1}{\boldsymbol{\beta}}_i{x}_{i-N}=\textbf{y}_j, 1\le{j}\le{M}\vspace{-0.1cm}
	\end{equation}
	
	The weights $\textbf{B}=(\boldsymbol{\beta}_1^T, \boldsymbol{\beta}_2^T,...,\boldsymbol{\beta}_N^T, \boldsymbol{\beta}_{N+1}^T,..., \boldsymbol{\beta}_{N+d_1}^T)^T$ do not only include the connection between enhancement nodes and output layer, but also include the connection between input layer and output layer (shown as Fig.~\ref{RVFL_structure}). With $d_1$ and $N$ nodes from the original and enhancement nodes respectively, there will be accordingly $N+d_1$ weights (i.e., $\boldsymbol{\beta}_i, i=1,..., N+d_1$ values), to be determined. The weights can be achieved by $\textbf{B}=\textbf{H}^+\textbf{Y}$ ($\textbf{H}$ is the matrix version of the features of all samples from enhancement nodes and input layer) in the similar way with ELM, improving the training efficiency significantly.

	\subsection{Broad Learning System}
	
\begin{figure}[ht!]
		\centering
		\includegraphics[scale=0.65]{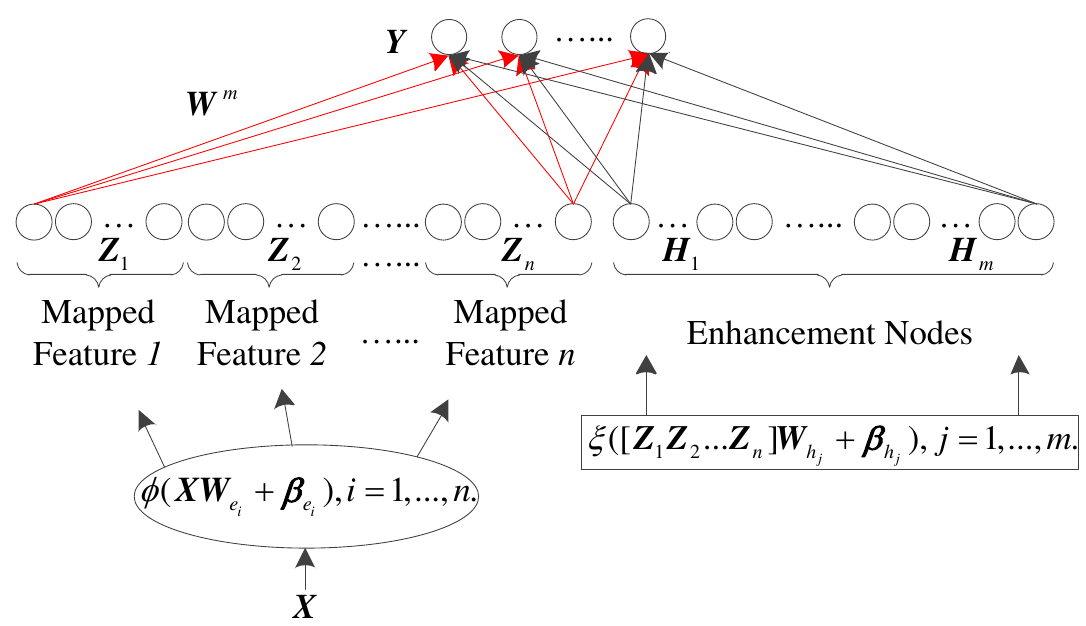}
		\caption{The architecture of broad learning system.}
		\label{BLS_Structure}
	\end{figure}

	Broad learning system (BLS) was proposed by Liu and Chen \cite{chen2018broad} recently, inspired by the architecture of the traditional RVFL neural network \cite{pao1992functional}. Unlike the RVFL that takes the inputs directly and establishes the enhancement nodes, the inputs in BLS are firstly converted into a set of random features (i.e., mapped features in Fig.~\ref{BLS_Structure}) by the given feature functions. Then these features are further connected to the enhancement by linear or nonlinear activation functions. Finally, the mapped features along with the enhancement nodes are directly fed to the output and the weights in the output layer can be learned by the efficient pseudo inverse approximation \cite{rao1971generalized}. The detail structure of BLS is illustrated in Fig.~\ref{BLS_Structure}.  
	
	Given the training data $(\textbf{x}_i,\textbf{y}_i)|\textbf{x}_i\in{\mathcal{R}^{M}},\textbf{x}_i\in{\mathcal{R}^{C}}, i=1,..., N$, with $n$ feature mappings $h_i, i=1,...,n$ and $m$ groups enhancement nodes, a typical BLS is constructed as follows.
	
	The $i$th mapped feature could be denoted as $\textbf{\textit{Z}}_i=h_i(\textbf{\textit{XW}}_{ei}+\boldsymbol{\beta}_{ei}), i=1,...,n$. The parameters $\textbf{\textit{XW}}_{ei}$ and the bias $\boldsymbol{\beta}_{ei}$ are generated from the given distribution with proper dimensions. With all $n$ groups of features $\textbf{\textit{Z}}^n\overset{\Delta}{=}[\textbf{\textit{Z}}_1,...,\textbf{\textit{Z}}_n]$, the $j$th enhancement nodes is illustrated as:	
	\begin{align}
		\label{BLS_1}
		\textbf{\textit{H}}_j=\xi_j(\textbf{\textit{Z}}^n\textbf{\textit{W}}_{hj}+\boldsymbol{\beta}_{hj}), j = 1,..., m
	\end{align}
	where $\xi_j$ is some prior activation function and the total enhancement nodes can be described as $\textbf{\textit{H}}^m\overset{\Delta}{=}[\textbf{\textit{H}}_1,...,\textbf{\textit{H}}_m]$. Accordingly, the approximation result of the above BLS can be represented as: 
	\begin{align}
		\label{BLS_2}
		\textbf{\textit{Y}}& =[\textbf{\textit{Z}}_1,\textbf{\textit{Z}}_2,...,\textbf{\textit{Z}}_n,\textbf{\textit{H}}_1,\textbf{\textit{H}}_2,...,\textbf{\textit{H}}_m]\textbf{\textit{W}}^m =[\textbf{\textit{Z}}^n,\textbf{\textit{H}}^m]\textbf{\textit{W}}^m
	\end{align} 	
	$\textbf{\textit{W}}^m$ can be calculated via $\textbf{\textit{W}}^m\overset{\Delta}{=}[\textbf{\textit{Z}}^n,\textbf{\textit{H}}^m]^{+}\textbf{\textit{Y}}$. In this way, BLS provides an alternative way of learning deep structure without time-consuming by avoiding abundant parameters in multiple layers and the fine-tune step based on the backpropagation. Other alternative versions of BLS are detailed in \cite{chen2018broad}.  
	
	\section{Related Work}\label{Related_work}
	
	Many existing works have demonstrated that a single model can seldom maintain good generalization across a variety of applications \cite{krawczyk2017ensemble, adhikari2015neural, song2017multi, donate2013time, chandra2006ensemble, li2016ensemble}. An ensemble of multiple models with a suitable combination rule leads to better predictive performance than that could be obtained from any of the constituent models alone \cite{zhang2017multiobjective}. Although an ensemble may perform better than the single model, it is not easy to construct a good ensemble model when faced with the crucial tasks arising from generating multiple members where each member can contribute to the constructed ensemble maximally, and a suitable combination rule design. 
	
	A variety of promising strategies have been developed for generating the members for ensemble modelling in TS prediction, consisting of heterogeneous component learners \cite{adhikari2015neural} and homogeneous base learners. The former includes the ensemble with more than one predictors while the latter consists of various architectures of the base learner \cite{qiu2014ensemble}, a variety of parameter combinations \cite{song2017multi}, TS decomposition \cite{qiu2017empirical} and different feature vectors \cite{wang2015hierarchical}. Base learners such as support vector machine (SVM) and ANN, are applied to generate individual forecasting models in \cite{adhikari2015neural}, where a linear combination is proposed to determine the combining weights after analyzing their patterns in successive in-sample forecasting trials. In \cite{lin2017random}, since the weights and biases of hidden neurons in ELM are all randomly generated, $M$ ELMs are used as the base learners. The bootstrap sampling technique is adopted to generate the training samples for every ELM, and then RF model is applied to aggregate these ELM models. The training outputs of deep learning network at different epochs (epoch=100, 200,..., 2000) are considered as the ensemble members and SVR is applied to combine the outputs from various deep belief networks (DBNs) as the ensemble in \cite{qiu2014ensemble}. TS is decomposed into several intrinsic mode functions (IMFs) via empirical mode decomposition and each component is trained with DBN \cite{qiu2017empirical}, then the prediction results of IMFs are combined via either unbiased or weighted summation to obtain the final ensemble. TS is decomposed with different levels of discrete wavelet transform (DWT), where each of the components is trained with ELM and combined using partial LS \cite{li2016ensemble}. In \cite{adhikari2012novel}, three TS forecasting models (i.e., ARIMA, ANNs and EANNs) are employed to build the ensemble and a nonlinear weighted ensemble mechanism is proposed to aggregate individuals by considering correlations among them. 
	
	To date, no studies demonstrate which method outperforms the rest. Feature selection techniques such as sequential forward selection (SFS) and sequential backward selection (SBS), are  popular methods used for selecting the optimal subset of models for the ensemble \cite{song2017multi}. Moreover, the aforementioned studies only consider the generation of ensemble members and ignore the relationship among them. When the correlation among members is high, the combined performance could be similar to that of each of the constituents. Therefore, diversity among all the members may provide another criterion to evaluate each individual \cite{chen2010multiobjective}, where negative correlation learning (NCL) is the popular method to evaluate the model diversity. By maintaining accuracy and diversity simultaneously, an EEL employs MOEA to obtain the members via the evolved PF \cite{chandra2006ensemble}. MOEA is applied to optimize the structure of (recurrent neural network) RNN and obtain the ensemble members from the evolved PF \cite{smith2014evolutionary}. Different selection methods have been used for pruning the members in PF for the final ensemble learning. However, this approach is only proposed for univariate TS prediction. Multi-objective optimization is applied to remaining useful life (RUL) estimation in prognostic with optimizing the structure of DBN, where the impact of different TWs as the parameters has been investigated on simulation data \cite{zhang2017multiobjective}. The superiority of the proposed MODBNE is only demonstrated on the RUL data and it does not consider selecting the influential channels of MTS. Previously, we proposed an evolutionary multi-objective ensemble learning (EMOEL) for electricity consumption prediction with various auxiliary factors \cite{song2018evolutionary}.	
	
	To the best of our knowledge, studies on the EEL for MTS prediction are relatively sparse. The proposed EEL framework mainly focuses on producing multiple PFs by MOEA to construct SEL for addressing MTS problems consisted of CS, FE, and predictive modelling. Since existing work only considers the best ensemble performance from one evolved PF without considering whether the remaining evolved PFs have good models that can further improve the ensemble performance, this is the first work that integrates models from the evolved PFs to build the SEL. The differences between the proposed EEL and the previously proposed EMOEL \cite{song2018evolutionary} are depicted as follows:	
	
	\begin{itemize}[leftmargin=*] 
		\item For each selected channel, instead of choosing one from all extracted features, each FE method is binary in EEL, which means it is possible to select the features from heterogeneous methods, including statistical methods (i.e., mean, maximum, minimum and standard deviation), DTW (i.e., 1-level, 2-level, 3-level and 4-level decomposition) and PLA (with three different parameters).
		\item In addition to ELM, RVFL and BLS are chosen as the base learners in EEL. Compared with ELM, RVFL has another parameter which is the connection between the input layer and output layer. The corresponding parameters in BLS are the number of windows of feature nodes, feature nodes per window and enhancement nodes.
		\item Instead of exploring the performance under each population size, SEL is performed on the evolved PFs from multiple population sizes in EEL. 
		\item Multi-resolution based on different granularities of a TS is proposed in \cite{song2017multi}. However, the resolutions are constructed with certain statistics calculation from its associated time interval using one specific TW. EEL considers multiple resolutions on different TWs.
		\item The novel EEL is verified on two real-world MTS datasets.
	\end{itemize}

	\section{The Proposed Approach}\label{Proposed_method}
	
	\subsection{Framework}
\begin{figure}[ht!]
		\centering
		\includegraphics[scale=0.75]{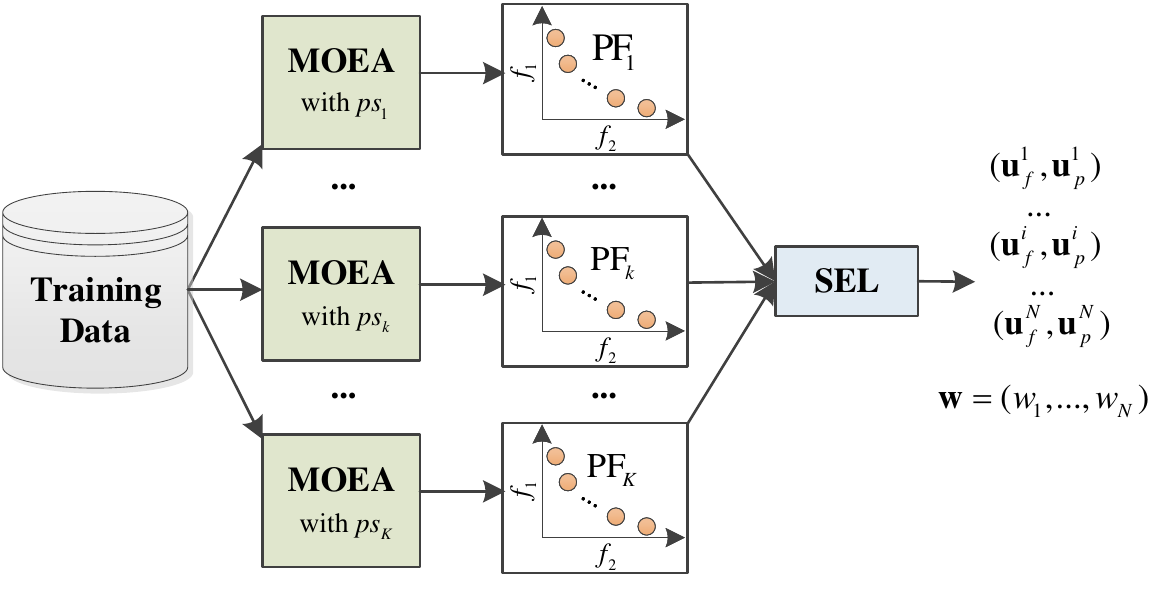}
		\caption{Illustration of the training process of the proposed EEL framework, where $\text{PF}_k$ denotes the set of Pareto front (PF) solutions (represented in Fig.~\ref{Dimension_Structure}) produced by an MOEA with population size $\textit{ps}_k$ and the selective ensemble learning (SEL) module selects an optimal subset from all $K$ sets of PF solutions and optimizes the combination coefficients (represented in Fig.~\ref{Testing_process}) to pursue the optimal ensemble prediction result.}
		\label{overall_framework}
	\end{figure}

	Given MTS data $\textbf{X} \in \mathbb{R}^{L\times (d+1)}, \textbf{X}=(\textbf{x}^1, \textbf{x}^2,..., \textbf{x}^{d+1})$, each component in $\textbf{X}$ is an ordered sequence of $L$ real-value observations and the $i^{th}$ time sequence is $\textbf{x}^i=(x_1^i, x_2^i,..., x_L^i)^T, i=1,..., d+1$. We mainly use the auxiliary MTS $\textbf{x}^i, i = 2, ..., d + 1$ and the historical information of $\textbf{x}^1$ to perform one-step ahead prediction on $\textbf{x}^1$. To address this problem, we propose a novel EEL framework, illustrated in Fig.~\ref{overall_framework}. 
	
	With the training dataset, $K$ MOEAs over different population sizes ($\textit{ps} = \{\textit{ps}_1, \textit{ps}_2,..., \textit{ps}_K\}$) are used to search for the optimal solutions consisted of CS ($\textbf{u}_{cs}=(u_{cs}^2, u_{cs}^{3},..., u_{cs}^{d}), u_{cs}^{n} \in {\{0,1}\}, n=2,...,d$), TW selection ($\textbf{u}_{tw}=(u_{tw}^1, u_{tw}^{2},..., u_{tw}^{d+1}), u_{tw}^i \in \mathcal{T}_i$, $\mathcal{T}_i$ represent the TWs of $i^{th}$ TS), resolutions $\text{u}_{r} \in \mathcal{R}$ \cite{song2017multi} for $\textbf{x}^1$, FE (i.e., perform FE or not, $\textbf{u}_{fe} = (u_{fe}^1, u_{fe}^{2},..., u_{fe}^{d}), \text{u}_{fe}^{i} \in {\{0,1}\}$) and FE method selection ($\textbf{u}_{fs} = (\textbf{u}_{fs}^1,\textbf{u}_{fs}^2,...,\textbf{u}_{fs}^{d})$), and PM relevant parameters ($\textbf{u}_p = ({u}_p^1, {u}_p^2,..., {u}_p^{d_p}), {u}_p^k \in \mathcal{P}_k, k = 1, ..., d_p$, $\mathcal{P}_k$ is the $k^{th}$ parameter settings) by considering the prediction accuracy (error) and model correlation (diversity) as two conflicting objectives (i.e., $(f_1, f_2)$ in Fig.~\ref{overall_framework}). Each eventually evolved MOEA leads to a PF composed of non-dominated trained models and accordingly $K$ PFs are generated. The number of generated ensemble models in the $k^{th}$ PF (i.e., $\text{PF}_k, k=1, 2,..., K$) is in relation to $\textit{ps}_k$. Afterwards, SEL is performed on the generated PFs to select the optimal subset of models and then the selected models are linearly combined with LS regressor. The outputs of SEL include CS and FE relevant parameters $\textbf{u}_f^i, i = 1,..., N$, PM relevant parameters $\textbf{u}_p^i$, and ensemble coefficients $\textbf{w} = (w_1,..,w_N)$.

	\textbf{Objective Functions:} The decision variables consists of CS and FE parameters $\textbf{u}_f=(\textbf{u}_{tw},\text{u}_{r},\textbf{u}_{cs},\textbf{u}_{fe}, \textbf{u}_{fs})$ involved in CS and FE in the pipeline and PM parameters $\textbf{u}_p$. With TS data $\textbf{X}$, the accuracy and diversity functions (i.e., $f_1(\textbf{X}, \textbf{u}_{f}, \textbf{u}_{p})$ and $f_2(\textbf{X}, \textbf{u}_{f}, \textbf{u}_{p})$) are optimized by an MOEA. 
	
	Accuracy: To maximize the accuracy of an ensemble member means to minimize the prediction error, evaluated by Root Mean Square Error (RMSE) as Eq.~\ref{1objective}: 
	\begin{align}
		\text{Minimize:} \quad f_1^j(\textbf{y},\hat{\textbf{y}}_j)=\sqrt{\frac{1}{s}\sum\limits_{i=1}^{s}(y^i-\hat{y}_j^i)^2}, \nonumber\\
		\hat{\textbf{y}}_j=f(\textbf{X}, \textbf{u}_f^j, \textbf{u}_p^j), j=1,...,M
		\label{1objective}
	\end{align}
	$y^i$ ($i= 1,...s, s \text{ is the total number of the training samples}$) is the real value of $i^{th}$ training sample and $\hat{y}_j^i$ represents the estimated prediction result obtained by $j^{th}$ predictor for the $i^{th}$ training sample.
	
	Diversity: To maximize the diversity between the outputs of different ensemble members is to minimize the correlation. NCL \cite{chandra2006ensemble} is used to define the diversity as follows:
	\begin{align}
		\text{Minimize:}\quad f_2^j(\hat{\textbf{y}}_j,\hat{\textbf{Y}})=\sum\limits_{i=1}^{s}(\hat{y}_j^i-\bar{y}^i)\sum\limits_{m\neq j, m=1}^{M}(\hat{y}_m^i-\bar{y}^i) \nonumber\\
		\hat{\textbf{y}}_j=f(\textbf{X}, \textbf{u}_f^j, \textbf{u}_p^j), j=1,...,M;
		\bar{{y}}^i=\frac{1}{M}\sum\limits_{j=1}^{M}\hat{{y}}_j^i
		\label{2objective}
	\end{align}
	where $\hat{\textbf{Y}}=(\hat{\textbf{y}}_1, \hat{\textbf{y}}_2,..., \hat{\textbf{y}}_M)$, $\hat{y}_j^i$ and $\hat{y}_m^i$ represent the outputs of the $j^{th}$ and $m^{th}$ base learner for the $i^{th}$ training sample, respectively. $\bar{y}^i$ is the average output of all base predictors on $i^{th}$ training sample. 
	
	\begin{figure}[h!]
		\centering	
		\includegraphics[scale=0.65]{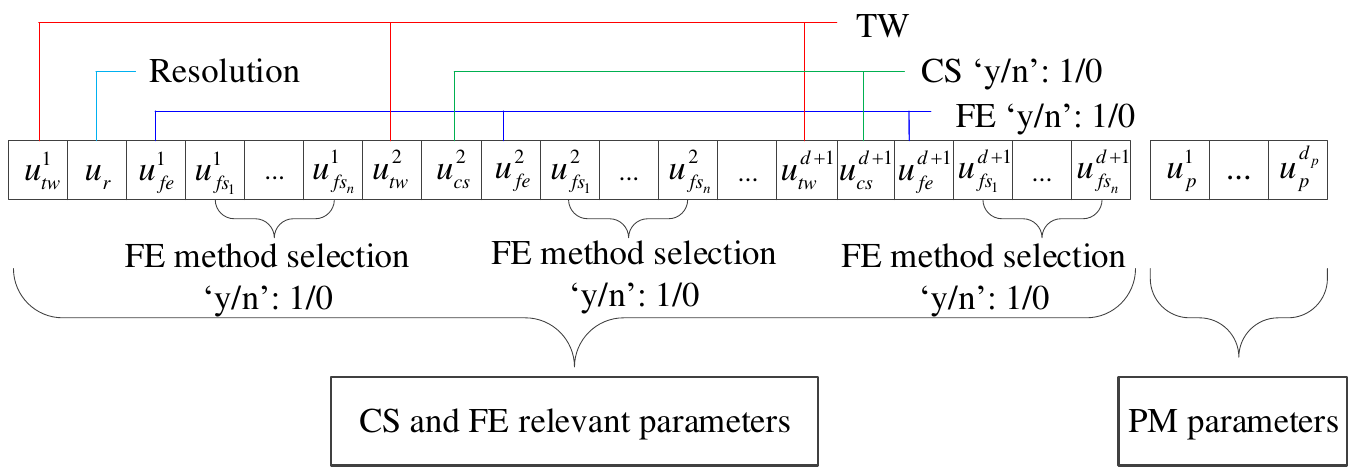}
		
		\caption{Solution representation in MOEA, where channel selection (CS) denotes whether the corresponding channel is selected or not, resolution represents the time series sampling interval \cite{song2017multi} for the channel to be predicted, feature extraction (FE) represents whether FE is applied to the time series segment with its length denoted by time window (TW) w.r.t. a selected channel, FE method selection represents whether a specific feature extraction method is applied or not, and prediction model (PM) parameters represent the parameters in the predictor.}
		\label{Dimension_Structure}
	\end{figure}
	
	\begin{algorithm*}[h!]
		\footnotesize
		\caption{MOEA/D integrated with the base learner (i.e., RVFL, ELM and BLS)}	\label{MOEAD}
		\SetAlgoLined
		\KwIn{A MOP with its solution space $\Omega$ and F objectives, $M$ (also $\textit{ps}$): population size (i.e. subproblems in MOEA/D), $T$: neighborhood size, $\textit{maxFEs}$: maximal number of evaluations in MOEA/D}
		\KwOut{Pareto front solutions, each solution corresponds to an ensemble member}
		\textbf{Step 1) Initialization:}\\
		\textbf{Step 1.1)} Randomly generate an initial population $\{\textbf{x}_1,..., \textbf{x}_M\}$ with uniform distribution in search space $\Omega$, then for each individual  $\textbf{x}_i, i=1,..., M$, generate the training dataset $i$ by decoding each dimension into its real space, $\#\textit{FEs}=0$ \\
		\textbf{Step 1.2)} Evaluate each candidate solution $\textbf{x}_i, i=1,..., M$ via the $i^{th}$ generated training dataset and $i^{th}$ base leaner to obtain its objective fitness $\textbf{Z}(\text{x}_i)$, $\#\textit{FEs}=\#\textit{FEs}+\textit{ps}$\\   
		\textbf{Step 1.3)} Randomly generate $M$ uniformly distributed weights
		$\{\lambda_1,..., \lambda_M\}$, where $\lambda_i=\lambda^1,..., \lambda^F$ with respect to $i^{th}$ individual,  its $T$ closest weight vectors $\{\lambda_{i_1},..., \lambda_{i_T}\}$ are obtained to form its neighborhoods $B(i)=\{{i_1},..., {i_T}\}$\\
		\textbf{Step 1.4)} Initial reference point $\textbf{Z}^*=\{z^{1^*},..., z^{F^*}\}$, satisfied 
		$z^{f^*}=\min\limits_{i=\{1,..., M\}} z^f(\textbf{x}_i), f=1,..., F$\\
		\textbf{Step 1.5)} Initialize normalization factors $\tilde{\textbf{Z}}=\{\tilde{z}^1,..., \tilde{z}^f\}$, satisfied $\tilde{z}^f=\max\limits_{i=\{1,..., M\}} |z^f(\textbf{x}_i)|, f=1,..., F$	\\
		\textbf{Step 2) Evolution:}\\
		\While{$\textit{FEs} < \textit{maxFEs}$}{
			\textbf{Step 2.1) Adaptive normalization:}In the current population, for each individual, apply adaptive normalization to its fitness on each objective, i.e., $z^f(\textbf{x}_i)=\frac{z^f(\textbf{x}_i)}{\tilde{z}^f}, i=1,..., M, f=1,..., F$\\
			\For {$i=1,..., M$}{
				\textbf{Step 2.2) Reproduction:} Randomly generate two indexes $k$, $l$ from $B(i)$, and then generate a new solution $i'$ from $\textbf{x}_k$ and $\textbf{x}_l$ by using the DE operator along with a Gaussian mutation applied under probability of 0.5: 
				\begin{align*}
					\textbf{x}_{i'}=\!\left\{\begin{array}{rl}
						\!\!\!\textbf{x}_i+0.5\cdot(\textbf{x}_k-\textbf{x}_l)+rnd(0,\sigma) & \mbox{if}\quad rnd(0,\sigma)\leq 0.5 \\ 
						\textbf{x}_i+0.5\cdot (\textbf{x}_k-\textbf{x}_l) & \mbox{otherwise}  
					\end{array}\right.
				\end{align*} 
				Each element in $\sigma$ is set to one twentieth of the corresponding decision variable's range\\
				\textbf{Step 2.3) Repairing:} Apply a problem-specific repair on the newly generated $\textbf{x}_{i'}$ to limit each of its elements in lower (upper) bound \\
				\textbf{Step 2.4) Evaluation:} Decode the newly generated $\textbf{x}_{i'}$ into its real space (i.e., indexes for feature parameters and parameters in the base learner), generate $i'^{th}$ training dataset according to the new feature parameters, and evaluate  $\textbf{x}_{i'}$ via $i'^{th}$ base leaner using corresponding training dataset and obtain $\textbf{Z(x)}_{i'}$, $\#\textit{FEs}=\#\textit{FEs}+1$\\
				\textbf{Step 2.5) Adaptive normalization:} Normalize $\textbf{x}_{i'}$, i.e., $z^f(\textbf{x}_{i'})=\frac{z^f(\textbf{x}_{i'})}{\tilde{z}^f}, i=1,..., M, f=1,..., F$ \\ 
				\textbf{Step 2.6) Replacement:} For each $i_s \in B(i)$, if $\max\limits_{f\in\{1,..., F\}}\lambda^f_{i_s}\cdot|z^f(\textbf{x}_{i'})-z^{f^*}|\leq \max\limits_{f\in\{1,..., F\}}\lambda^f_{i_s}\cdot|z^f(\textbf{x}_{i_s})-z^{f^*}|$, set $\textbf{x}_{i_s}=\textbf{x}_{i'}$ and $z^f(\textbf{x}_{i_s})=z^f(\textbf{x}_{i'})$\\
				\textbf{Step 2.7) Update reference point:} If $z^f(\textbf{x}_{i'})<z^{f^*}$, set $z^{f^*}=z^f(\textbf{x}_{i'})$
			}
			\textbf{Step 2.8) Update normalization factors:} $\tilde{\textbf{Z}}=\{\tilde{z}^1,..., \tilde{z}^f\}$, satisfied $\tilde{z}^f=\max\limits_{i=\{1,..., M\}} |z^f(\textbf{x}_i)|, f=1,..., F$
		}						
	\end{algorithm*}

	\textbf{Encoding and Decoding in MOEA:} The decision variables in MOEA include TWs $\textbf{u}_{tw}$, resolution $u_r$, CS $\textbf{u}_{cs}$ for each TS in $\textbf{X}$, FE $\textbf{u}_{fe}$ for each TS in $D$, FE method selection $\textbf{u}_{fs}$ and PM parameter set $\textbf{u}_{p}$, all of which are binary or discrete integer values when fed to the PM. Each variable in $(\textbf{u}_{tw},\text{u}_{r},\textbf{u}_{cs},\textbf{u}_{fe}, \textbf{u}_{fs}, \textbf{u}_p)$ is encoded to the range of $[0,1]$ when initializing or updating the candidate solutions. When evaluating each solution, each dimension of the solution is decoded to its real range. TWs $\textbf{u}_{tw}=(u_{tw}^1, u_{tw}^{2},..., u_{tw}^{d+1}), \text{u}_{tw}^{i} \in \mathcal{T}, i=1,...,d+1$ ($\mathcal{T}_i$: TWs for the $i$th TS, e.g., if the range of the TWs for the $i^{th}$ TS is from 2 to 10 with interval 2, $\mathcal{T}_i=\{1, 2, 3, 4, 5\}$ represent the corresponding TWs), the respective dimensions that represent $\textbf{u}_{tw}$ are decoded to $\mathcal{T}, \mathcal{T}=\{\mathcal{T}_1, \mathcal{T}_2,...,\mathcal{T}_{d+1}\}$ to find the real TWs used. The dimensions corresponding to resolution $u_r$ and PM parameter set $\textbf{u}_{p}$ are also decoded to their real values by considering $\mathcal{R}$ and $\mathcal{P}, \mathcal{P}=\{\mathcal{P}_1,...,\mathcal{P}_{d_p}\}$, respectively. For CS $\textbf{u}_{cs}=(\text{u}_{cs}^2, ..., \text{u}_{cs}^{d+1})$, FE (1: perform FE; 0: raw data) $\textbf{u}_{fe}=(\text{u}_{fe}^1, ..., \text{u}_{fe}^{d+1})$ and FE method selection $\textbf{u}_{fs}=(\textbf{u}_{fs}^1,...,\textbf{u}_{fs}^{d+1}), \textbf{u}_{fs}^i=(\text{u}_{{fs}_1}^i,...,\text{u}_{{fs}_n}^i)$ (each TS has $n$ features extracted, i.e., $\{mean$, $maximum$, $minimum$, $std$, ${DTW}_1$, ${DTW}_2$, ${DTW}_3$, ${DTW}_4$, ${PLA}_1$, ${PLA}_2$, ${PLA}_3\}$ and each feature is binary, shown as in Fig.~\ref{Dimension_Structure}), the respective dimensions for each solution are decoded to 0 (not selected) or 1 (selected). Instead of applying one FE method to MTS, considering the independent characteristics among MTS and the superiority of heterogeneous features extracted from multiple FE methods, each selected TS may choose more than one features from these $n$ features, as illustrated in Fig.~\ref{Dimension_Structure}. 
	\begin{figure}[h!]
		\centering
		\includegraphics[scale = 0.8]{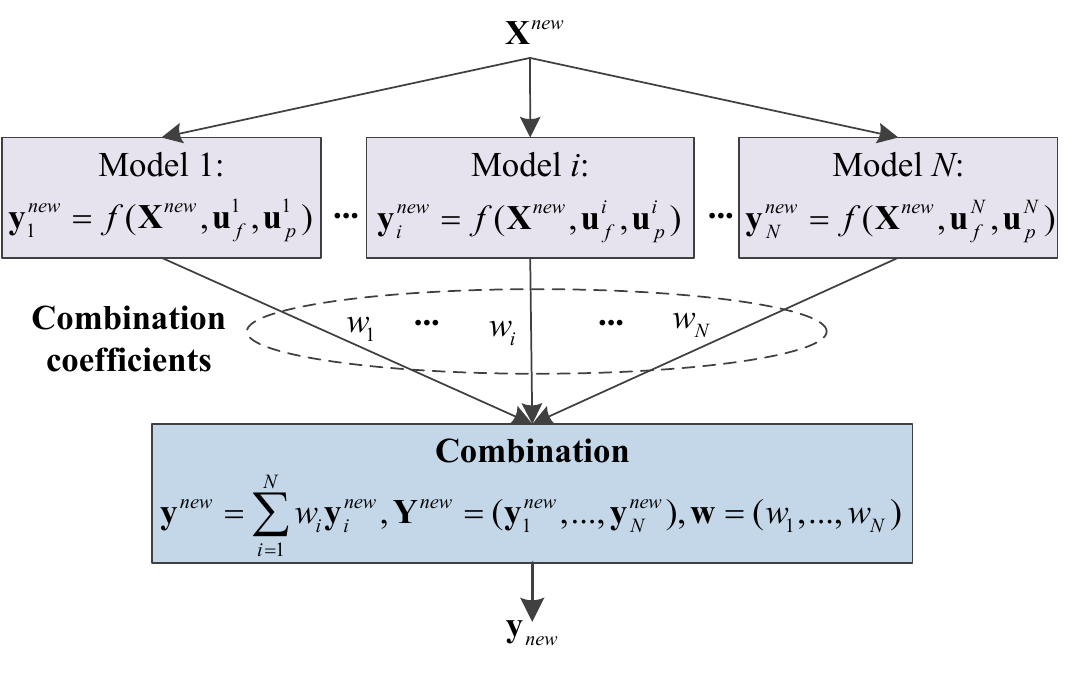}
		\caption{Application of the trained model via EEL to predict the output $\textbf{y}^{\textit{new}}$ of a new input $\textbf{X}^{\textit{new}}$.}
		\label{Testing_process}
	\end{figure}
	
	\textbf{The testing procedure:} During the training procedure, the optimal subset of models (i.e., solutions) over PFs can be obtained via SEL. Given new input data $\textbf{X}^{\textit{new}}$ to test the proposed EEL, the steps are illustrated in Fig.~\ref{Testing_process}. With the obtained $\textbf{u}_f$, $\textbf{u}_p$, and inputs $\textbf{X}^{\textit{new}}$, the predicted $\textbf{y}^{\textit{new}}_i, i = 1, 2,..., N$ over the $i^{th}$ selected model can be obtained via $f(\textbf{X}^{\textit{new}}, \textbf{u}_f^i, \textbf{u}_p^i)$ so that $N$ predicted instances $\textbf{Y}^{\textit{new}} = (\textbf{y}^{\textit{new}}_1, \textbf{y}^{\textit{new}}_2,..., \textbf{y}^{\textit{new}}_N)$ can be obtained. $\textbf{Y}^{\textit{new}} = (\textbf{y}^{\textit{new}}_1, \textbf{y}^{\textit{new}}_2, ...,  \textbf{y}^{\textit{new}}_N)$ are combined with combination coefficients $\textbf{w}=(w_1, w_2,..., w_N)$ to obtain the ensemble result. The ensemble learning result $\textbf{y}^{\textit{new}}$ is obtained via $\textbf{y}^{\textit{new}}= \sum\limits_{i=1}^{N} w_i\textbf{y}^{\textit{new}}_i,\sum\limits_{i=1}^{N}w_i=1$.

	\subsection{Implementation}
	
	RVFL, ELM and BLS are three neural networks with random weights which do not have many parameters to be learned. The number of hidden neurons is the only parameter in ELM. Compared with ELM, RVFL has another parameter which is the connection between the input layer and output layer. The corresponding parameters in BLS are the number of windows of feature nodes, feature nodes per window and enhancement nodes. Due to the super-fast computational ability and promising performance, RVFL, ELM and BLS are easier to be applied to regression and classification problems \cite{tang2018non, tang2016extreme, wan2014probabilistic, chen2018broad} and accordingly they are chosen as the base learners in EEL. Considering the time-consuming problem caused by optimizing RVFL, ELM and BLS simultaneously in one MOEA/D, they are optimized individually. Then the implementation of ensemble members generation with MOEA/D and SEL are described for addressing the problem presented.
	
	\subsubsection{Multi-objective Optimization Algorithm}
	
	MOEA/D \cite{zhang2007moea} has achieved great success in the field of MOPs and has attracted extensive attention. MOEA/D explicitly decomposes a MOP into $M$ scalar optimization subproblems. These subproblems are to be addressed simultaneously by involving a population of solutions. At each generation, the population is composed of the best solution found so far for each subproblem. Each subproblem has a uniformly distributed vector as its weight. By calculating Euclidean distance between two weights, each subproblem has $T$ closest neighborhoods. Therefore, the optimal solutions to two neighborhoods is similar. Each subproblem is solved by only using its neighborhoods' information, which makes it much more efficient in solving MOP compared with other MOEAs. EEL employs MOEA/D to find the solutions for the decision variables $\textbf{u}=(\textbf{u}_f,\textbf{u}_p)$ which consists of CS and FE relevant parameters and base learner (RVFL, ELM or BLS) relevant parameters. The details of MOEA/D with the base leaner are depicted in Algorithm.~\ref{MOEAD}. 
	
	As the objective fitnesses have different magnitudes, where some magnitudes are much larger than others. To overcome the problems caused by undesirable bias during the search direction, we apply adaptive normalization to each of the objective fitness $z^f, f=1,..., F$ ($F$ is the number of objectives and $F$=2 in this task, $z^f=(f_1, f_2)$) as follows:
	\begin{align}
		z^f(\textbf{x})=\frac{z^f(\textbf{x})}{\tilde{z}^f}
	\end{align}   
	where $\tilde{z}^f$ is the normalization factor updated at beginning of every generation, obtained by $\tilde{z}^f=\max\limits_{i=\{1,..., M\}} |z^f(\textbf{x}_i)|$.
	
	\subsubsection{Ensemble Methods}
	
	Mean (the coefficient for each ensemble model equals to $\frac{1}{M}$) and LS are two popular methods for ensemble learning, but they do not remove the redundant and noisy models. To obtain the optimal subset of the ensemble models that can lead to the best prediction accuracy, two SEL methods are designed, where SFS and SBS are applied to select the optimal subset of models and LS (denoted as $\text{SFS+LS}$ and $\text{SBS+LS}$) is used to learn the coefficients to combine the selected models. $\text{SFS+LS}$ starts with the best solution in the ensemble models and adds the one with the best performance among the remaining models for each iteration. $\text{SBS+LS}$ mainly focuses on removing one model whose removal can lead to higher accuracy for each iteration. Since $\text{SBS+LS}$ works similarly to $\text{SFS+LS}$, we only present the implementation of $\text{SFS+LS}$ performed on the models of PFs to select the instances $\hat{\textbf{Y}}_s$ and the respective index set $s_{id}$ and obtain the combination coefficients $\textbf{w}$ in Algorithm.~\ref{SFS_Algorithm}.
	
	\begin{algorithm}[h!]
		\footnotesize
		\caption{The implementation of $\text{SFS+LS}$}
		\SetAlgoLined
		\label{SFS_Algorithm}
		\KwIn{$\hat{\textbf{Y}}=(\hat{\textbf{Y}}_1,\hat{\textbf{Y}}_2,...,\hat{\textbf{Y}}_K)$; $acc=+\infty$; $\textbf{y}$; $\hat{\textbf{Y}}_s=\emptyset$; $s_{id}$; \tcp{$\textbf{y}$: Real values}}
		\KwOut{$\hat{\textbf{Y}}_s$; $acc$, $\textbf{w}$, $s_{id}$}
		Remove the same elements in $\hat{\textbf{Y}}$ as $\hat{\textbf{Y}}=(\hat{\textbf{y}}_1, \hat{\textbf{y}}_2,...)$\\
		\tcp{SFS+LS performed on $\hat{\textbf{Y}}$}
		\While{$\sim\text{isempty}(\hat{\textbf{Y}})$}{$id=\emptyset$; \\
			\For{$k=1:length(\hat{\textbf{Y}})$}{
				\nonl ${\textbf{T}}=\hat{\textbf{Y}}_s\cup\hat{\textbf{y}}_{k}$\\
				\nonl $\textbf{w}=\textbf{T}^+ \textbf{y}$; $\hat{\textbf{y}}=\textbf{T} \textbf{w}$ \tcp{Solved by LS, $\hat{\textbf{y}}$: predicted values}
				\nonl $acc_\textit{new}=\text{RMSE}(\textbf{y}, \hat{\textbf{y}})$\tcp{Evaluation}
				\If{$acc_\text{new}<acc$}{
					\nonl $acc=acc_\textit{new}; id=k; flag=1$}}
			\If{$\sim\text{isempty}(id)$}{
				\nonl $\hat{\textbf{Y}}_s=\hat{\textbf{Y}}_s\cup\hat{\textbf{y}}_{id}$;\tcp{Add $\hat{\textbf{y}}_{id}$  to $\hat{\textbf{Y}}_s$}
				\nonl $s_{id}=s_{id}\cup{id}$;\tcp{Add ${id}$ to the selected index set $s_{id}$}
				\nonl $\hat{\textbf{Y}}=\hat{\textbf{Y}} \setminus \hat{\textbf{y}}_{id}$\tcp{Remove $\hat{\textbf{y}}_{id}$ from $\hat{\textbf{Y}}$}
				\Else{
					\nonl\textbf{break}}
		}}
		\Return $\hat{\textbf{Y}}_s$; $acc$, $\textbf{w}$, $s_{id}$
	\end{algorithm} 
	
	\begin{table}[h!]
		\centering
		\footnotesize
		\caption{The parameter settings in the proposed EEL}
		\begin{tabular}{|c|p{6.8cm}|}
			\hline
			EEL    & \qquad \qquad \qquad Parameter settings: Dataset A \\\hline
			MOEA/D &  {\raggedright Neighborhood size $T=4$; the maximal fitness evaluations $\textit{MaxFEs}=2.5e4$; $\textit{ps}=\{30, 50, 80, 100, 120, 150$\};  $\mathcal{T}_1=\{1, 2,..., 16\}, \mathcal{R}=\{1, 2,..., 15\}, \mathcal{T}_i=\{1, 2,..., 22\}, i=2,...,d+1, d=5$, where $\mathcal{T}_1$ represents the index set of the TW settings (6 to 96 with interval 6) for the prediction target; $\mathcal{R}$ is the index set of resolution settings for prediction target; $\mathcal{T}_i$ describes the index set of the TW settings (6 to 48 with interval 2) for the $i^{th}$ auxiliary TS; for each TS, the number of features $n=11$}                  \\
			\hline
			RVFL   &  {\raggedright ${d}_p=2$; $D=86$; $\mathcal{P}_1=\{1, 2,..., 40\}$: the index set of the hidden neurons settings (10 to 400 with interval 10); $\mathcal{P}_2=\{0, 1\}$: if there is connection between input and output layers or not}                \\
			\hline
			ELM    &  {\raggedright ${d}_p=1$; $D=85$; $\mathcal{P}_1=\{1, 2,..., 40\}$ (10 to 400 with interval 10): the index set of hidden neuron settings}                  \\
			\hline
			BLS    &   {\raggedright ${d}_p=3$; $D=87$; $\mathcal{P}_1=\{1, 2,..., 20\}$: the index set of the number of windows of feature nodes (1 to 20 with interval 1); $\mathcal{P}_2=\{1, 2,..., 50\}$: feature nodes per window (1 to 50 with interval 1); $\mathcal{P}_3=\{1, 2,..., 150\}$: the number of enhancement nodes (10 to 1500 with interval 10)}                   \\\hline
			EEL    & \qquad \qquad \qquad Parameter settings: Dataset B \\\hline
			MOEA/D &  {\raggedright  $\mathcal{T}_1=\{1, 2,..., 12\}, \mathcal{R}=\{1, 2,..., 15\}, \mathcal{T}_i=\{1, 2,..., 12\}, i=2,...,d+1, d=5$ (the values in $\mathcal{T}_i, i=1, 2,...,d+1$ represent the TW from 2 to 24 with interval 2)}                  \\
			\hline
			RVFL   &  {\raggedright $\mathcal{P}_1=\{1, 2,..., 40\}$ and $\mathcal{P}_2=\{0, 1\}$}                \\
			\hline
			ELM    &  {\raggedright $\mathcal{P}_1=\{1, 2,..., 30\}$ (10 to 300 with interval 10)}                  \\
			\hline
			BLS    &   {\raggedright $\mathcal{P}_3=\{1, 2,..., 100\}$: 10 to 1000 with interval 10}                   \\\hline
		\end{tabular}
		\label{Parameter_setting_EEL}
	\end{table}
	\begin{table}[h!]
		\centering
		\footnotesize
		\caption{The parameter settings in the state-of-the-art models}
		\begin{tabular}{|c|p{6.5cm}|}
			\hline
			\begin{tabular}[c]{@{}c@{}}Comparison \\ models\end{tabular} & \qquad \qquad \qquad \qquad Parameter settings   \\\hline
			RVFL      &   {\raggedright Number of hidden neurons: 10 to 400 with interval 10, random generation of the connection between input layer and output layer} \\\hline
			ELM       & {\raggedright Number of hidden neurons: 10 to 400 and 300 with interval 10 for Dataset A and Dataset B, respectively}               \\\hline
			SVR        &  {\raggedright Gamma: $2^{-24}$ to $2^{15}$ with interval $2^{3}$; Cost: $2^{-10}$ to $2^{10}$ with interval $2^{3}$; kernel function: RBF}   \\\hline
			BLS         &   {\raggedright Number of windows: 2 to 20 with interval 5; feature nodes per window: 5 to 50 with interval 10; enhancement nodes: 50 to 1000 with interval 50 for Dataset A and 50 to 1500 with interval 50 for Dataset B}  \\\hline
			HELM        &  {\raggedright $N_1$: 10 to 300 with interval 30; $N_2$: 10 to 300 with interval 30; $N_3$: 10 to 300 with interval 30}              \\\hline
			LSTM       &   {\raggedright Number of hidden units: 5 to 100 with interval 5; dropout: 0.2, 0.5, 0.9, 0.99; learning rate: 0.001, 0.005, 0.01, 0.03, 0.05; batch size: 10, 20, 50, 100, 200}          \\\hline
			CNN & Learning rate: 0.001, 0.003, 0.005, 0.009, 0.01; CONV1: $f = 3$, $n = \{2, 4, 6, 8, 10\}$, $p$ = VALID, $s$ = 1; CONV2: $f = 2$, $n = \{4, 8, 12, 16, 20\}$, $p$ = VALID, $s$ = 1; POOL1: $f = 2$, $p$ = SAME, $s$ = 1; POOL2: $f = 2$, $p$ = SAME, $s$ = 1; number of neurons in fully connected layer: 5 to 50 with interval 5\\\hline
		\end{tabular}
		\label{Parameter_setting_compared}
	\end{table}
	\begin{table*}[h!]
		\centering
		\footnotesize
		\caption{Average RMSE performance of $\text{EEL}_\text{RVFL}$, $\text{EEL}_\text{ELM}$, and $\text{EEL}_\text{BLS}$ on $ps=$ 30, 50, 80, 100, 120, 150 under Mean, LS, $\text{SBS+LS}$, and $\text{SFS+LS}$ for Datasets A and B}
		\begin{tabular}{|c|c|ccc|ccc|}
			\hline
			\multicolumn{2}{|c|}{\multirow{2}{*}{}} & \multicolumn{3}{c|}{Dataset A}                       & \multicolumn{3}{c|}{Dataset B}                    \\\cline{3-8}
			\multicolumn{2}{|c|}{}    & $\text{EEL}_\text{RVFL}$             & $\text{EEL}_\text{ELM}$             & $\text{EEL}_\text{BLS}$            & $\text{EEL}_\text{RVFL}$           & $\text{EEL}_\text{ELM}$            & $\text{EEL}_\text{BLS}$            \\ \hline
			\multirow{6}{*}{Mean}       & 30      & 6.9822          & 5.6975          & 6.1340          & 166.14         & 146.31         & 51.08          \\
			& 50      & 7.7744          & 6.4598          & 6.6092          & 160.18         & 122.90         & 49.24          \\
			& 80      & 7.0321          & 6.0754          & 6.2562          & 93.12          & 71.61          & 50.17          \\
			& 100     & 7.6704          & 5.9806          & 6.4463          & 77.41          & 116.88         & 52.48          \\
			& 120     & 7.1425          & 5.8022          & 6.6295          & 93.10          & 85.24          & 50.90          \\
			& 150     & 7.0225          & 5.7892          & 6.5965          & 300.73         & 99.04          & 50.18          \\\hline
			\multirow{6}{*}{LS}         & 30      & 1.5284          & 1.4805          & 1.3487          & 731.21         & 182.55         & 21.61          \\
			& 50      & 1.4513          & 1.4780          & 1.3474          & 630.74         & 704.85         & 21.43          \\
			& 80      & 1.4645          & 1.5827          & 1.3443          & 691.16         & 301.10         & 20.97          \\
			& 100     & 1.4305          & 1.4228          & 1.3024          & 316.10         & 442.76         & 20.99          \\
			& 120     & 1.4796          & 1.4341          & 1.2797          & 440.53         & 576.36         & 20.89          \\
			& 150     & 1.3965          & 1.4327          & 1.2472          & 1079.51        & 548.36         & 21.25          \\\hline
			\multirow{6}{*}{SBS+LS}     & 30      & 1.4338          & 1.4709          & 1.3498          & 141.58         & 122.92         & 21.36          \\
			& 50      & 1.4461          & 1.4830          & 1.3388          & 368.22         & 107.74         & 21.00          \\
			& 80      & 1.4675          & 1.4757          & 1.3290          & 128.08         & 95.20          & 20.61          \\
			& 100     & 1.4260          & 1.4164          & 1.2851          & 217.31         & 119.61         & 20.43          \\
			& 120     & 1.4713          & 1.4301          & 1.2721          & 171.55         & 95.79          & 20.40          \\
			& 150     & 1.3978          & 1.4086          & 1.2223          & 397.81         & 216.55         & 20.20          \\\hline
			\multirow{6}{*}{SFS+LS}     & 30      & 1.4387          & 1.4620          & 1.3511          & 29.06          & 22.31          & 21.37          \\
			& 50      & 1.4444          & 1.5254          & 1.3406          & 24.07          & 23.11          & 20.98          \\
			& 80      & 1.4499          & 1.5004          & 1.3278          & 21.76          & 21.88          & 20.63          \\
			& 100     & 1.4242          & \textbf{1.4080} & 1.2820          & 21.99          & \textbf{20.73} & 20.32          \\
			& 120     & 1.4573          & 1.4092          & 1.2704          & \textbf{21.47} & 21.11          & 20.28          \\
			& 150     & \textbf{1.3772} & 1.4114          & \textbf{1.2219} & 25.82          & 21.53   & \textbf{20.15}\\\hline
		\end{tabular}
		\label{Inidividual_ps}
	\end{table*}	
	\section{Experiments}\label{Experiments}
	
	To demonstrate the superior performance of EEL over one real-world electricity consumption dataset and one benchmark PM2.5 Data of Five Chinese Cities Data Set \cite{liang2016pm2} from the UCI repository, the following hypotheses are verified: 
	
	\begin{itemize}
		
		\item The SEL performed on multiple PFs plays a significant role in boosting the prediction accuracy. 
		
		\item The proposed EEL outperforms the state-of-the-art models on two real-world MTS prediction tasks.
	\end{itemize}

	\subsection{Data Description}
	\subsubsection{Dataset A: Electricity Data Set}
	
	The data includes historical electricity data\footnote{https://www.rmit.edu.au/about/our-values/sustainability/sustainable-urban-precincts-program}\textsuperscript{,}\footnote{https://github.com/hsong60csmile/electricity-sample-data} and environmental factors (temperature, dew point, humanity, wind speed and sea level). Electricity data is collected from the smart meters of the buildings in the city campus of RMIT University, Melbourne, at 15-minute intervals. The environmental factors are crawled from an online weather station\footnote{https://www.wunderground.com/} that broadcasts periodic readings every 20 minutes. The dataset used is from 01 October 2017 to 31 December 2017 in Melbourne, Australia. There are 7392 samples in all, which are randomly created into training (occupy 2/3) and testing (occupy 1/3) datasets five times.

	\subsubsection{Dataset B: PM2.5 Data of Five Chinese Cities Data Set}
	
	The PM2.5 Data of Five Chinese Cities Data Set\footnote{https://archive.ics.uci.edu/ml/datasets/PM2.5+Data+of+Five+Chinese+Cities} consists of date information, PM2.5 concentration ($ug/m^3$), dew point, humidity, temperature, pressure, cumulated wind speed (m/s), among others, and includes PM2.5 data on 5 different cities. Each of them is an hourly reading from 01 January 2010 to 31 December 2015. There is much empty data for PM2.5 concentration before 2014 and hence, data from 01 January 2015 to 31 December 2015 on Beijing is used. The total of 8400 samples are randomly assigned to training (occupy 2/3) and testing (occupy 1/3) datasets five times.
	
	\begin{figure*}[h!]\centering
		\subfigure[$\text{EEL}_\text{RVFL}$]
		{\centering\scalebox{0.37}
			{\includegraphics{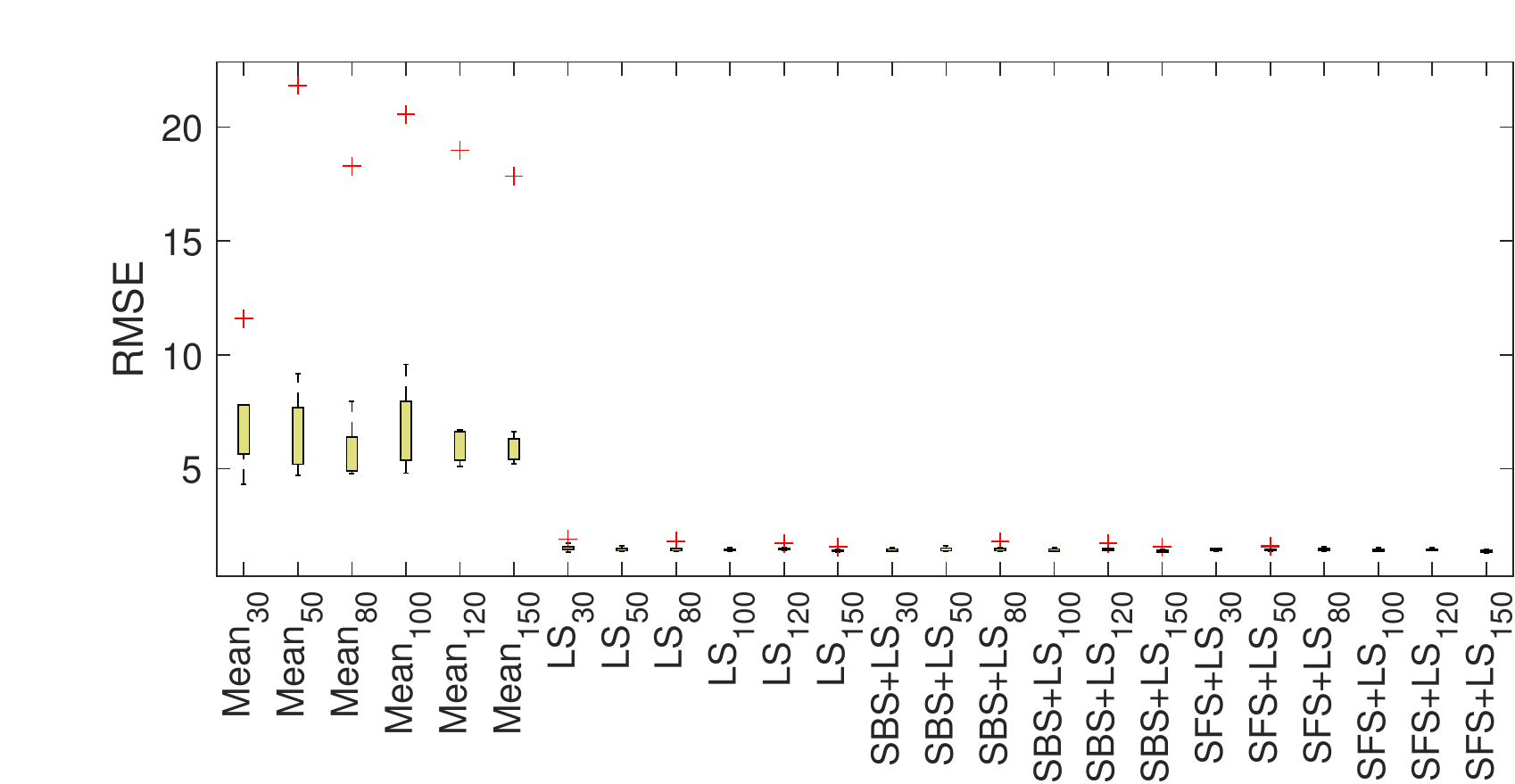}}\label{Ele_RVFL}}
		\subfigure[$\text{EEL}_\text{ELM}$]
		{\centering\scalebox{0.37}
			{\includegraphics{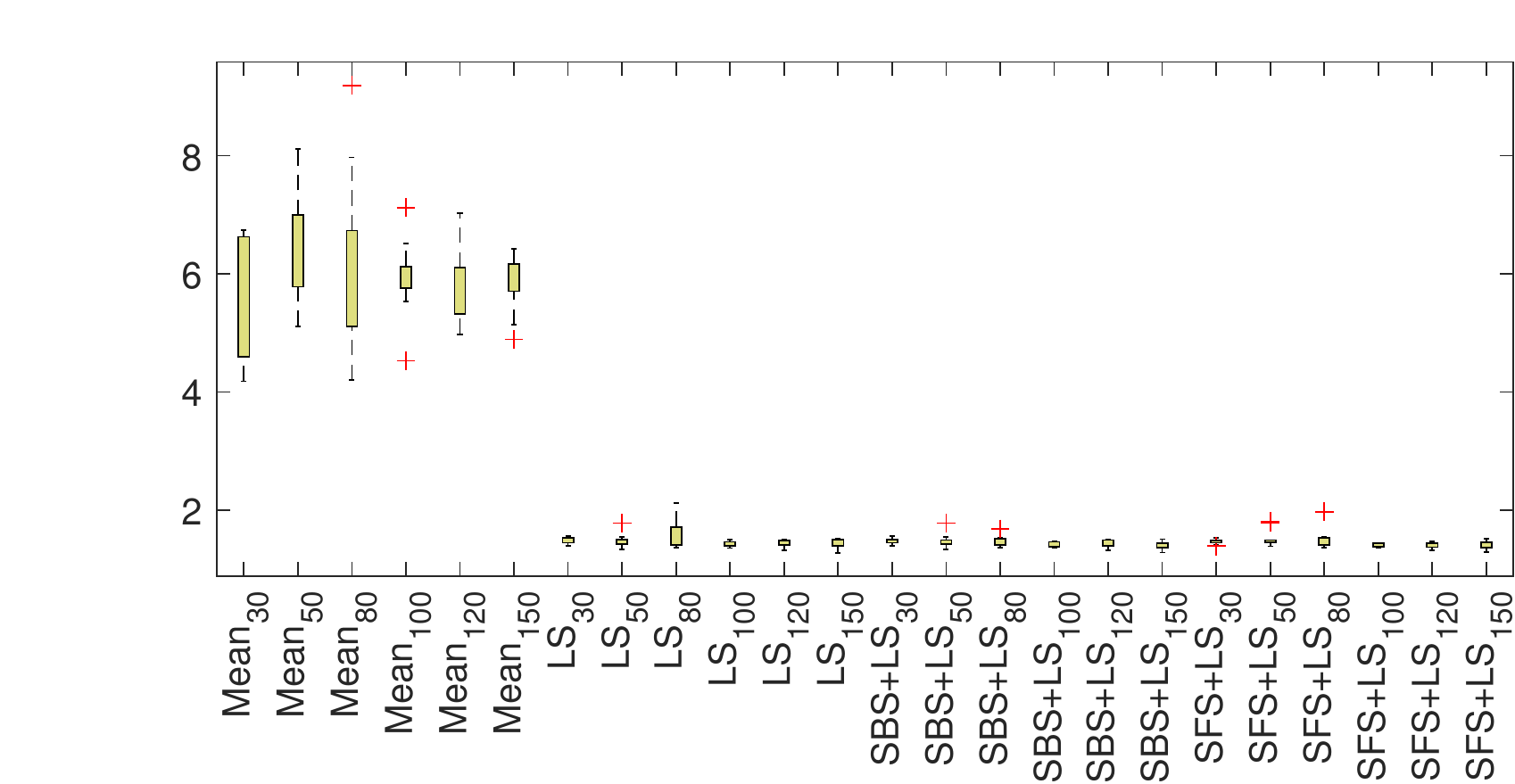}}\label{Ele_ELM}}
		\subfigure[$\text{EEL}_\text{BLS}$]
		{\centering\scalebox{0.37}
			{\includegraphics{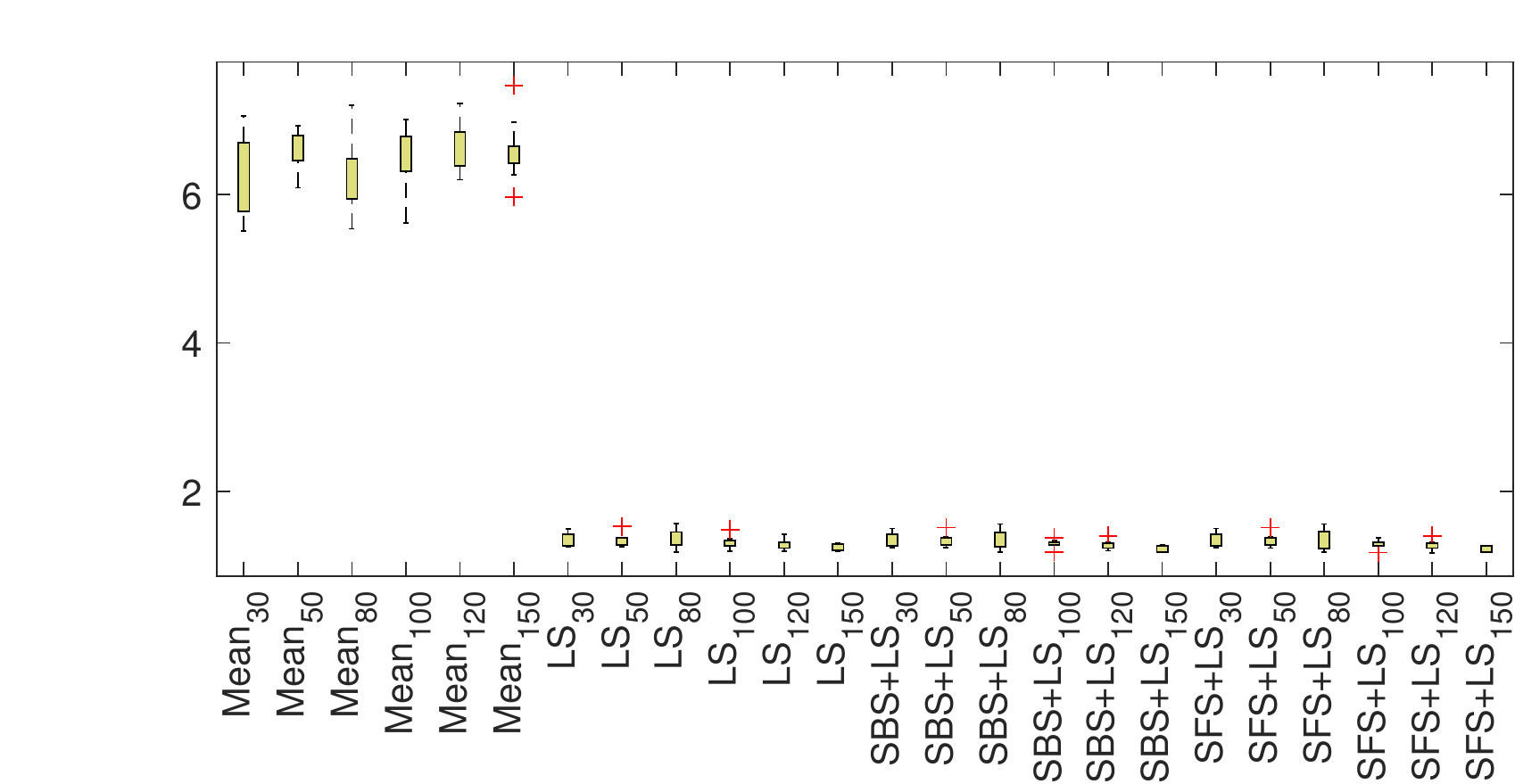}}\label{Ele_BLS}}
		\begin{minipage}{18cm}~\\
			\scriptsize Note: \textcolor{red}{+} in the box plot represent outliers
		\end{minipage}
		\caption{Box plots of EEL using RVFL, ELM and BLS as base learners with $ps = 30, 50, 80, 100, 120, 150$ over different ensemble models for Dataset A.} \label{individula_DatasetA}
	\end{figure*}

	\begin{figure*}[h!]\centering
		\subfigure[$\text{EEL}_\text{RVFL}$]
		{\centering\scalebox{0.37}
			{\includegraphics{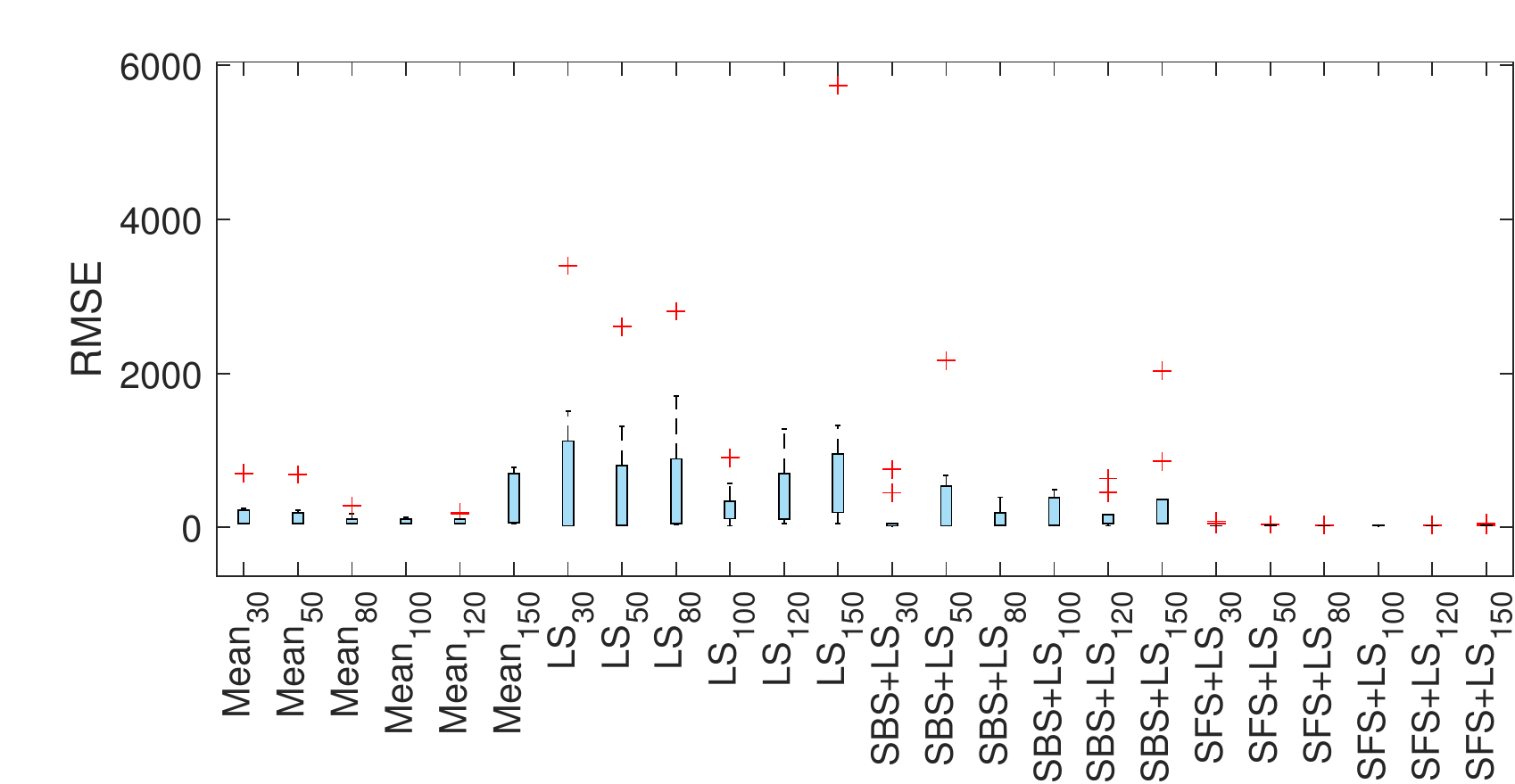}}\label{Air_RVFL}}
		\subfigure[$\text{EEL}_\text{ELM}$]
		{\centering\scalebox{0.37}
			{\includegraphics{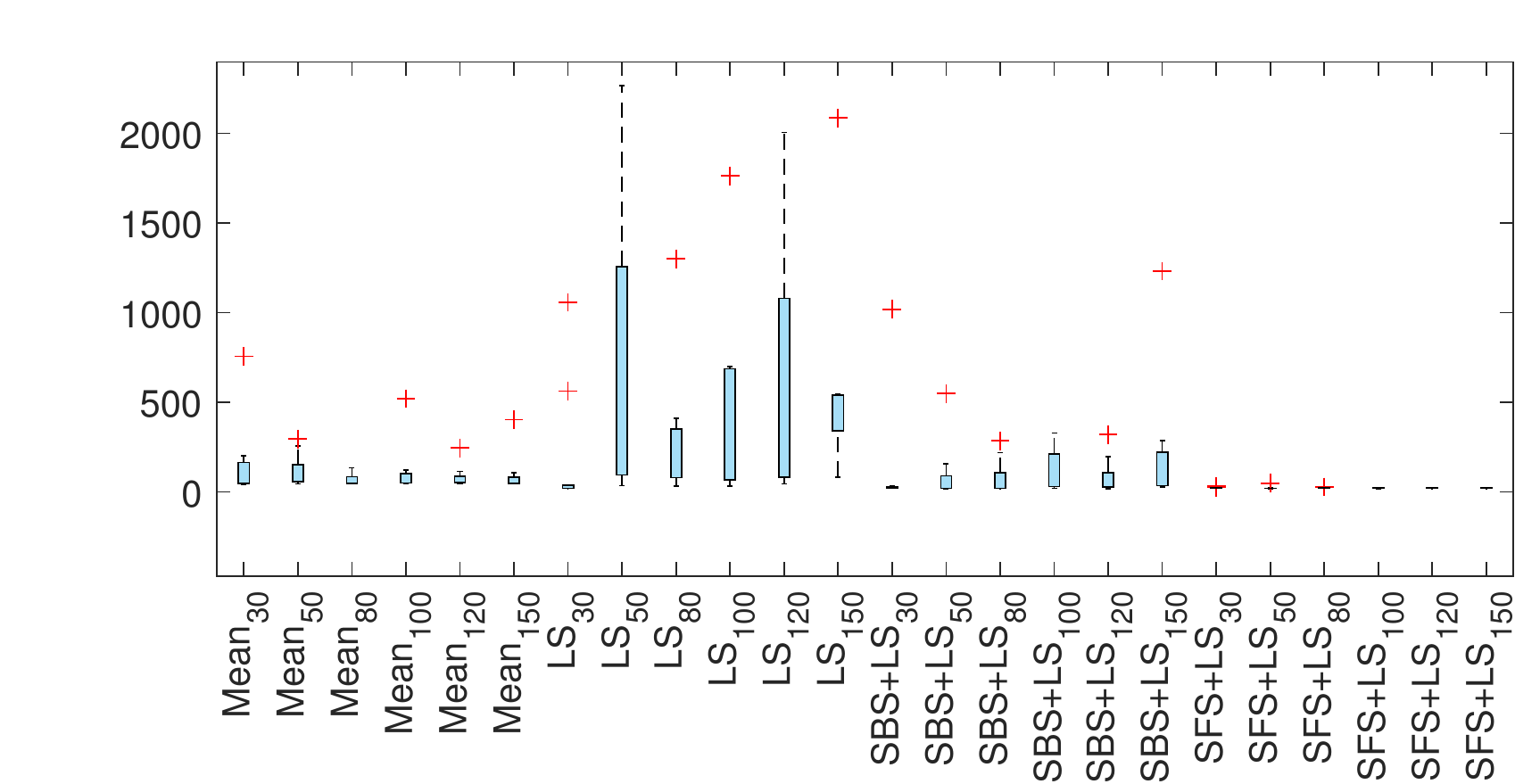}}\label{Air_ELM}}
		\subfigure[$\text{EEL}_\text{BLS}$]
		{\centering\scalebox{0.37}
			{\includegraphics{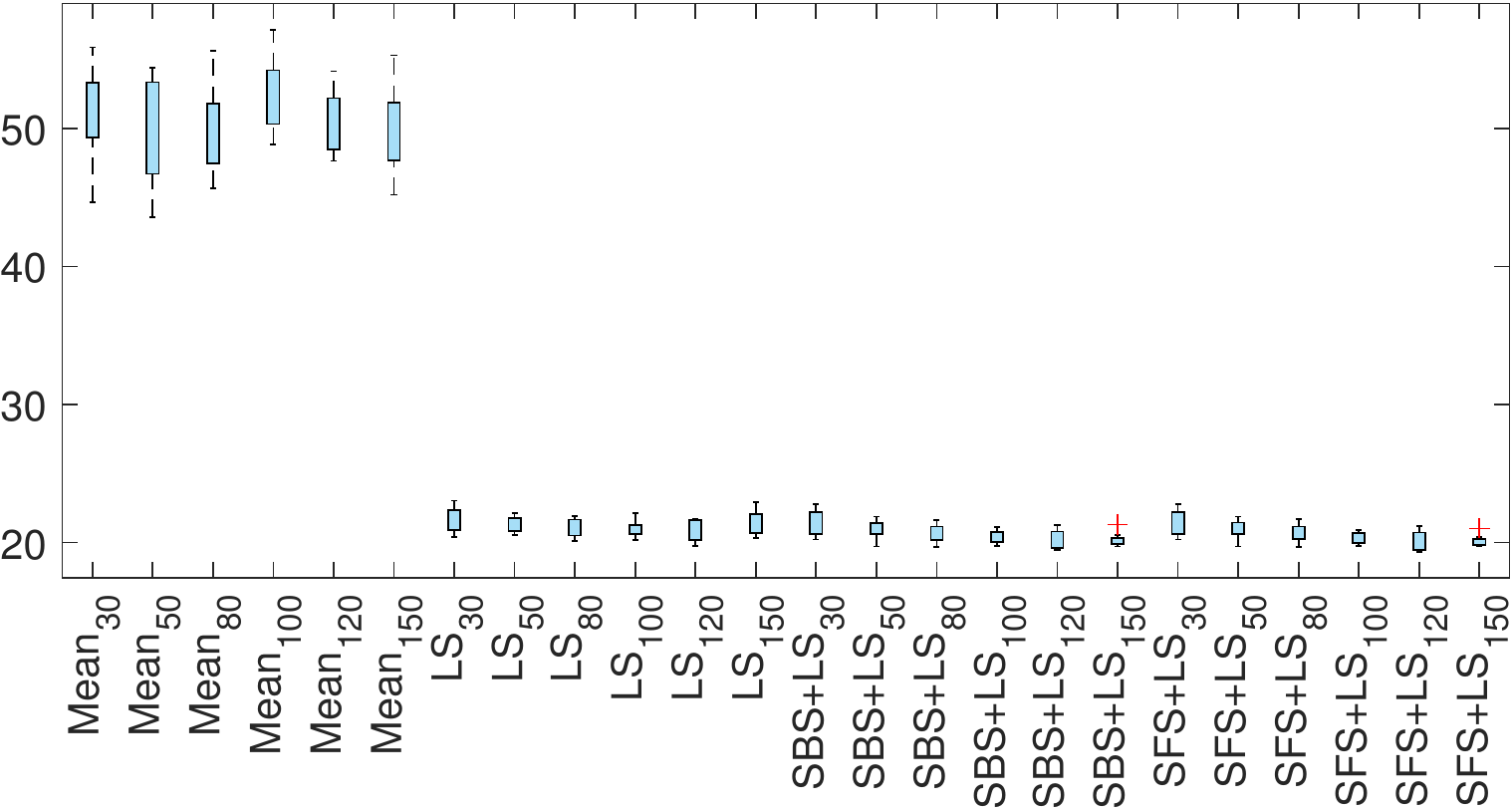}}\label{Air_BLS}}
		\caption{Box plots of EEL using RVFL, ELM and BLS as base learners with $ps = 30, 50, 80, 100, 120, 150$ over different ensemble models for Dataset B.} \label{individula_DatasetB}
	\end{figure*}
	
	\subsection{Experimental Setup}
	The experimental setup for the proposed EEL includes the parameters of the model and the range of decision variables. The performance of EEL is investigated based on three PMs (i.e., RVFL, ELM and BLS), where multiple PFs are generated from different population sizes in MOEA/D. The parameter settings in EEL for each dataset are presented in Tab.~\ref{Parameter_setting_EEL} (If not specified for Dataset B, they are same as the settings in Dataset A).  
	
	The result comparison includes several state-of-the-art models such as ELM, RVFL, SVR \cite{chang2011libsvm}, BLS, Hierarchical ELM (HELM) \cite{tang2016extreme}, LSTM \cite{gers1999learning}, convolutional neural networks (CNNs) \cite{lecun1995convolutional, krizhevsky2012imagenet}, and dual-stage attention-based recurrent neural network (DA-RNN) \cite{qin2017dual}, where HELM, LSTM, CNN, and DA-RNN are the models with the ability to learn the features. LSTM is one of the popular deep learning models applied in solving TS prediction problems given its ability to capture long-term dependencies \cite{karevan2020transductive}. CNN can extract useful patterns from multiple dimensions and accordingly, it has been considered to address TS prediction problems \cite{koprinska2018convolutional}. DA-RNN is one of the popular attention-based MTS PMs. We apply DA-RNN to the datasets investigated using the parameters suggested in the original paper. We also tune the important parameters on our datasets and present the best results for comparison. For other comparison models, except for TW settings (i.e., 6 to 96 with interval 6 for Dataset A and 2 to 24 with interval 2 for Dataset B), the auxiliary parameters are described in Tab.~\ref{Parameter_setting_compared} (If not specified, the parameter settings are the same for each dataset). The difference of $\mathcal{T}_1$ and $\mathcal{T}_i$ in Tab.~\ref{Parameter_setting_EEL} are based on the original TS data we have.

	\subsection{Result}
	
	The performance of EEL is comprehensively studied by choosing RVFL, ELM and BLS (i.e., $\text{EEL}_\text{RVFL}$, $\text{EEL}_\text{ELM}$ and $\text{EEL}_\text{BLS}$) as the base learners under different population sizes (i.e., $\textit{ps}$) in MOEA/D for both Datasets A and B, where two selective ensemble methods $\text{SBS+LS}$ and $\text{SFS+LS}$ are designed and performed on the generated PFs except Mean and LS methods. To demonstrate the effectiveness of multiple PFs in the ensemble modelling, the ensemble performed on each single PF is explored and compared. To distinguish the result over single PF and multiple PFs, $\text{EEL}^\Delta$ represents the results over multiple PFs. Finally, in comparison to several state-of-the-art models, the superiority of the proposed novel EEL is further verified. Each of the results presented in the tables below is the average RMSE over 10 runs.

	\subsubsection{Comparison with The Ensemble Modelling Performed on Single PF}
	
	The population size denotes the individuals generated from MOEA/D for the ensemble and accordingly $\textit{ps}$ is comprehensively studied. To investigate the ensemble performance on single PF generated by each of $\textit{ps}$ = 30, 50, 80, 100, 120, 150 using MOEA/D, the combination strategies such as Mean, LS, SBS+LS and SFS+LS are used to explore the prediction performance under the base learners RVFL, ELM and BLS for both Datasets A and B. Tab.~\ref{Inidividual_ps} reports the average RMSE over 10 runs for each of $\text{EEL}_\text{RVFL}$, $\text{EEL}_\text{ELM}$, and $\text{EEL}_\text{ELM}$ under different population sizes and the combination rules on both datasets. Mean and LS are directly applied to each evolved PF for the ensemble. $\text{SBS}+\text{LS}$ and $\text{SFS}+\text{LS}$ are SEL where SBS and SFS are applied to select the optimal subset of models from each evolved PF and LS is used to combine the selected models for the ensemble. The best average RMSE of EEL under each base learner across all ensemble rules and $ps$ is labelled in bold.
	
	\begin{table}[h!]
		\centering
		\footnotesize
		\caption{Comparison of EEL over multiple PFs ($\text{EEL}^\Delta$) and the best result over the single PF ($\text{EEL}^*$) of each base learner on Datasets A and B}
		\begin{tabular}{|p{1.2cm}|p{0.6cm}p{0.6cm}p{0.6cm}|p{0.6cm}p{0.6cm}p{0.6cm}|}
			\hline
			\multirow{2}{*}{\quad Models} & \multicolumn{3}{c|}{Dataset A}                       & \multicolumn{3}{c|}{Dataset B}                    \\\cline{2-7}
			& RVFL            & ELM             & BLS             & RVFL           & ELM            & BLS            \\\hline
			$\text{EEL}^*$ & 1.3772          & 1.4080          & 1.2219          & 21.47          & 20.73          & 20.15          \\
			$\text{EEL}_\text{LS}^\Delta$ & 1.4297          & 1.5220          & 1.2615          & 1894.79        & 894.55         & 21.81          \\
			$\text{EEL}_\text{SBS+LS}^\Delta$ & 1.3247          & 1.3325          & \textit{\textbf{1.1429}}          & 121.18         & 88.52          & \textit{\textbf{19.06}}          \\
			$\text{EEL}_\text{SFS+LS}^\Delta$ & \textbf{1.3127} & \textbf{1.3091} & \textit{\textbf{1.1384}} & \textbf{19.70} & \textbf{19.63} & \textit{\textbf{18.96}} \\\hline
		\end{tabular}
		\label{comparison_EEL}
	\end{table}
	
	\begin{table*}[h!]
		\centering
		\caption{Summarization of the number of individuals over all $ps$ settings and the number of individuals in $\text{EEL}_\text{LS}^\Delta$, $\text{EEL}_\text{SBS+LS}^\Delta$ and $\text{EEL}_\text{SFS+LS}^\Delta$ across each run for Datasets A and B}
		\begin{tabular}{|c|c|c|ccc|ccc|}
			\thickhline
			\multirow{2}{*}{Base learners} & \multirow{2}{*}{Total} & \multirow{2}{*}{Runs} & \multicolumn{3}{c|}{Dataset A}                                                 & \multicolumn{3}{c|}{Dataset B}   \\ \cline{4-9}
			&                        &                       & \multicolumn{1}{c}{$\text{EEL}_\text{LS}^\Delta$} & \multicolumn{1}{c}{$\text{EEL}_\text{SBS+LS}^\Delta$} & \multicolumn{1}{c|}{$\text{EEL}_\text{SFS+LS}^\Delta$} & \multicolumn{1}{c}{$\text{EEL}_\text{LS}^\Delta$} & \multicolumn{1}{c}{$\text{EEL}_\text{SBS+LS}^\Delta$} & \multicolumn{1}{c|}{$\text{EEL}_\text{SFS+LS}^\Delta$} \\ \thickhline
			\multirow{10}{*}{RVFL}         & \multirow{10}{*}{\begin{tabular}[c]{@{}c@{}}536\\ ($\sum\limits_{i=1}^{k}(ps_i+1)$)\end{tabular}}     & 1                     & 111                      & 69                       & 45                       & 225                      & 166                      & 42                       \\  
			&                        & 2                     & 119                      & 49                       & 36                       & 212                      & 161                      & 57                       \\ 
			&                        & 3                     & 101                      & 55                       & 29                       & 217                      & 116                      & 54                       \\  
			&                        & 4                     & 140                      & 59                       & 57                       & 212                      & 144                      & 55                       \\  
			&                        & 5                     & 133                      & 65                       & 34                       & 204                      & 113                      & 57                       \\  
			&                        & 6                     & 102                      & 63                       & 48                       & 229                      & 168                      & 57                       \\  
			&                        & 7                     & 122                      & 42                       & 40                       & 218                      & 127                      & 65                       \\ 
			&                        & 8                     & 158                      & 78                       & 65                       & 207                      & 95                       & 39                       \\  
			&                        & 9                     & 124                      & 64                       & 36                       & 228                      & 134                      & 66                       \\  
			&                        & 10                    & 125                      & 54                       & 42                       & 193                      & 123                      & 46                       \\\thickhline 
			\multirow{10}{*}{ELM}          & \multirow{10}{*}{\begin{tabular}[c]{@{}c@{}}536\\ ($\sum\limits_{i=1}^{k}(ps_i+1)$)\end{tabular}}     & 1                     & 115                      & 63                       & 39                       & 219                      & 130                      & 43                       \\ 
			&                        & 2                     & 100                      & 51                       & 42                       & 214                      & 152                      & 71                       \\ 
			&                        & 3                     & 143                      & 86                       & 62                       & 224                      & 126                      & 56                       \\ 
			&                        & 4                     & 187                      & 87                       & 81                       & 243                      & 186                      & 55                       \\ 
			&                        & 5                     & 124                      & 62                       & 46                       & 249                      & 149                      & 42                       \\ 
			&                        & 6                     & 144                      & 46                       & 44                       & 236                      & 159                      & 55                       \\ 
			&                        & 7                     & 176                      & 92                       & 59                       & 225                      & 166                      & 51                       \\ 
			&                        & 8                     & 121                      & 57                       & 33                       & 205                      & 161                      & 56                       \\ 
			&                        & 9                     & 103                      & 39                       & 26                       & 210                      & 127                      & 50                       \\  
			&                        & 10                    & 117                      & 66                       & 52                       & 209                      & 105                      & 66                       \\ \thickhline 
			\multirow{10}{*}{BLS}          & \multirow{10}{*}{\begin{tabular}[c]{@{}c@{}}536\\ ($\sum\limits_{i=1}^{k}(ps_i+1)$)\end{tabular}}     & 1                     & 229                      & 141                      & 62                       & 241                      & 90                       & 60                       \\ 
			&                        & 2                     & 223                      & 126                      & 60                       & 224                      & 70                       & 53                       \\ 
			&                        & 3                     & 235                      & 149                      & 83                       & 232                      & 79                       & 55                       \\ 
			&                        & 4                     & 202                      & 133                      & 57                       & 235                      & 94                       & 70                       \\ 
			&                        & 5                     & 244                      & 135                      & 73                       & 237                      & 77                       & 60                       \\ 
			&                        & 6                     & 245                      & 153                      & 82                       & 228                      & 95                       & 51                       \\ 
			&                        & 7                     & 233                      & 142                      & 77                       & 237                      & 90                       & 43                       \\  
			&                        & 8                     & 236                      & 125                      & 54                       & 228                      & 84                       & 61                       \\ 
			&                        & 9                     & 222                      & 127                      & 50                       & 246                      & 97                       & 63                       \\ 
			&                        & 10                    & 214                      & 131                      & 89                       & 237                      & 97                       & 61                       \\ \thickhline 
		\end{tabular}
		\label{Elec_Air_S}
	\end{table*}
	
	Tab.~\ref{Inidividual_ps} tells that Mean leads to the worst ensemble performance in comparison to LS, $\text{SBS}+\text{LS}$, and $\text{SFS}+\text{LS}$ over $\text{EEL}_\text{RVFL}$, $\text{EEL}_\text{ELM}$, and $\text{EEL}_\text{BLS}$ on Dataset A. For Dataset B, LS has worse prediction performance on $\text{EEL}_\text{RVFL}$ and $\text{EEL}_\text{ELM}$ while Mean has lowest accuracy on $\text{EEL}_\text{BLS}$. $\text{SBS}+\text{LS}$ and $\text{SFS}+\text{LS}$ lead to better performance in most cases compared with Mean and LS over all base learners for both datasets, which means using feature selection methods to select the optimal subset of individuals from the evolved PF for ensemble modelling improves the accuracy in comparison to combining all members together. Moreover, $\text{SFS}+\text{LS}$ outperforms other ensemble models such as Mean, LS, and $\text{SBS}+\text{LS}$ under each base learner for Datasets A and B. From the labelled bold values, we can observe that larger population sizes (i.e., $ps$ = 100, 120, 150) lead to the higher ensemble accuracy, which is influenced by the problem dimension and indicates that it is possible to improve the performance with a larger population size. Fig.~\ref{individula_DatasetA} and Fig.~\ref{individula_DatasetB} show box plots of RMSE from $\text{EEL}_\text{RVFL}$, $\text{EEL}_\text{ELM}$, and $\text{EEL}_\text{BLS}$ with $ps$ = 30, 50, 80, 100, 120, 150 obtained by Mean, LS, $\text{SBS} +\text{LS}$, and $\text{SFS} +\text{LS}$ over 10 runs for Datasets A and B, respectively. The findings corresponding to the results in Tab.~\ref{Inidividual_ps} are also evidenced by the RMSE distribution in Fig.~\ref{individula_DatasetA} and Fig.~\ref{individula_DatasetB}. 
	
	\begin{figure}[h!]\centering
		\subfigure[Dataset A]
		{\centering\scalebox{0.38}
			{\includegraphics{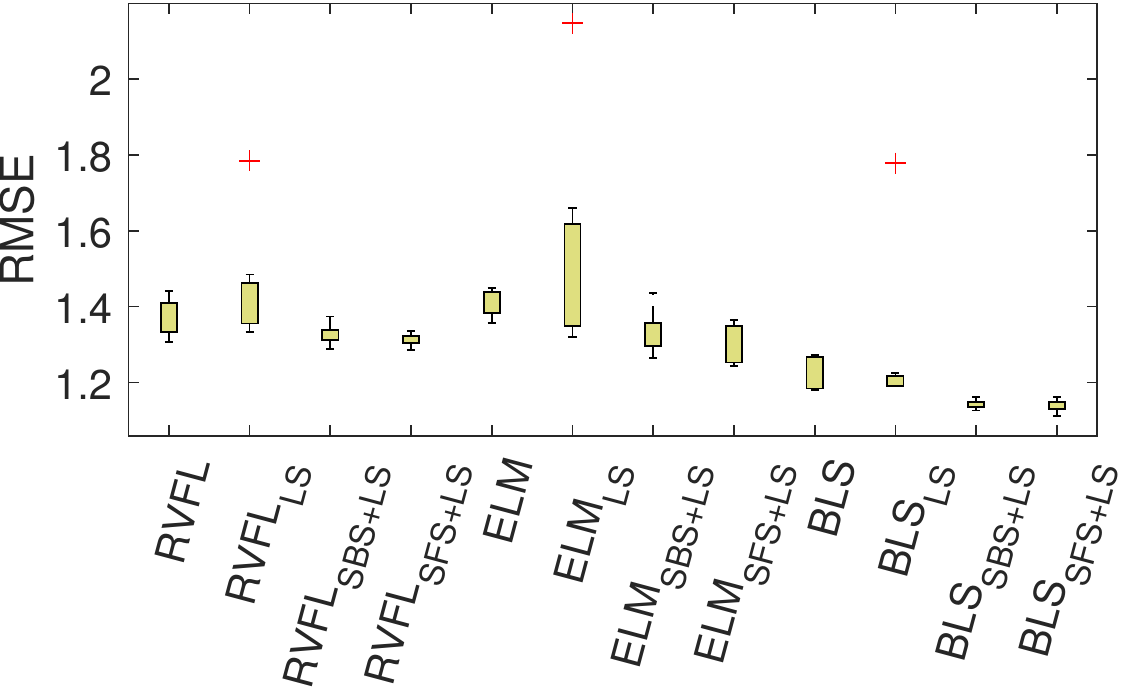}}\label{S_com_A}}\hspace{-0.1cm}
		\subfigure[Dataset B]
		{\centering\scalebox{0.38}
			{\includegraphics{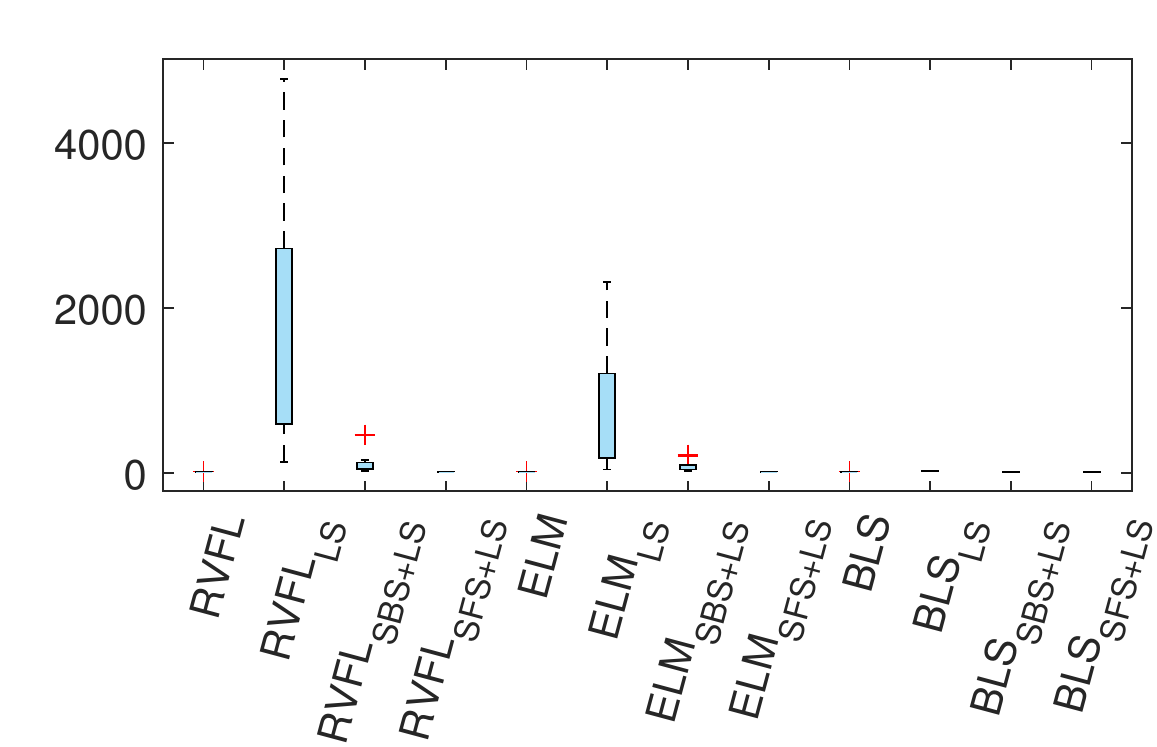}}\label{S_com_B}}\hspace{-0.0cm}
		\begin{minipage}{9cm}~\\
			\scriptsize Note: RVFL, ELM and BLS represent the results under $\text{EEL}^*$ and others denote the results under EEL with multiple PFs (i.e., $\text{EEL}^\Delta$). For example, RVFL represents $\text{EEL}^*$ over RVFL in Tab.~\ref{comparison_EEL}. $\text{RVFL}_\text{LS}$ represents $\text{EEL}_\text{LS}^\Delta$ under RVFL in Tab.~\ref{comparison_EEL}. 
		\end{minipage}	
		\caption{Box plots of EEL over multiple PFs and the best result of the single PF on RVFL, ELM and BLS for Datasets A and B.} \label{S_com}
	\end{figure}

	The performance of the proposed EEL is verified on three ensemble methods such as LS, $\text{SBS} +\text{LS}$, and $\text{SFS} +\text{LS}$ across each base learner. To verify the effectiveness of EEL over multiple PFs (denoted as $\text{EEL}^\Delta$), the highest prediction accuracy of $\text{EEL}_\text{RVEL}$, $\text{EEL}_\text{ELM}$, and $\text{EEL}_\text{BLS}$ from Tab.~\ref{Inidividual_ps} is used, denoted as $\text{EEL}^{*}$ over RVFL, ELM, and BLS. For each base learner, the best mean RMSE over $\text{EEL}^{*}$ and the $\text{EEL}^\Delta$ is labelled bold. The bold and italic values describe the results that show no difference from the lowest mean RMSE by using the Wilcoxon signed rank test over 10 runs. From the results reported in Tab.~\ref{comparison_EEL}, it is obvious to see that $\text{EEL}^\Delta$ outperforms $\text{EEL}^{*}$, which indicates that EEL with the multiple PFs leads to better prediction performance in comparison to using single PF. The superiority of $\text{EEL}^\Delta$ is further verified by the box plots in Fig.~\ref{S_com} through the smaller variance for Datasets A and B. 
	
	\begin{table*}[h!]
		\centering
		\footnotesize
		\caption{Comparison of RVFL, ELM, SVR, BLS, HELM, LSTM, CNN, DA-RNN \cite{qin2017dual}, $\text{EEL}_\text{RVFL}^\Delta$, $\text{EEL}_\text{ELM}^\Delta$, and $\text{EEL}_\text{BLS}^\Delta$ for Datasets A and B}
		\begin{tabular}{|c|cccccccc|ccc|}
			\hline
			& ELM    & RVFL   & SVR   & BLS    & HELM   & LSTM   & CNN & DA-RNN & $\text{EEL}_\text{RVFL}^\Delta$ & $\text{EEL}_\text{ELM}^\Delta$ & $\text{EEL}_\text{BLS}^\Delta$  \\\hline
			\begin{tabular}[c]{@{}c@{}}Dataset A\end{tabular} & 1.6712 & 1.6747 & 1.6681 & 5.0092 & 1.6547 & 1.4275 & 1.9142 & 2.0012 & \textbf{1.3127}    & \textbf{1.3091} & \textit{\textbf{1.1384}}   \\
			\begin{tabular}[c]{@{}c@{}}Dataset B\end{tabular} & 429.68 & 131.86 & 41.30  & 87.97  & 22.45  & 21.89 & 22.59 & 22.41 & \textit{\textbf{19.70}}     & \textit{\textbf{19.63}}   & \textit{\textbf{18.96}}    \\\hline
		\end{tabular}
		\label{all_com}
	\end{table*}
	
	The results also demonstrate that $\text{EEL}^\Delta$ with BLS leads to better ensemble performance than RVFL and ELM for both datasets. $\text{EEL}^\Delta$ with $\text{SBS}+\text{LS}$ (i.e., $\text{EEL}_\text{SBS+LS}^\Delta$) and $\text{SFS}+\text{LS}$ (i.e., $\text{EEL}_\text{SFS+LS}^\Delta$) have same ensemble accuracy, since their statistical test shows no significant difference on leading to the best ensemble performance. In comparison to combining all individuals in PFs using LS for ensemble modelling, SEL is a promising method given the accuracy obtained in Tab.~\ref{comparison_EEL}. The number of total individuals (Total) over all population sizes explored, the number of individuals in the PFs under each run, the number of individuals selected for each run over $\text{SBS+LS}$ and $\text{SFS+LS}$ are summarized in Tab.~\ref{Elec_Air_S} on both datasets, where the results further demonstrate the effectiveness of $\text{SBS}+\text{LS}$ and $\text{SFS}+\text{LS}$ from the number of individuals selected. 
	
	\begin{figure}[h!]\centering
		\subfigure[Dataset A]
		{\centering\scalebox{0.39}[0.39]
			{\includegraphics{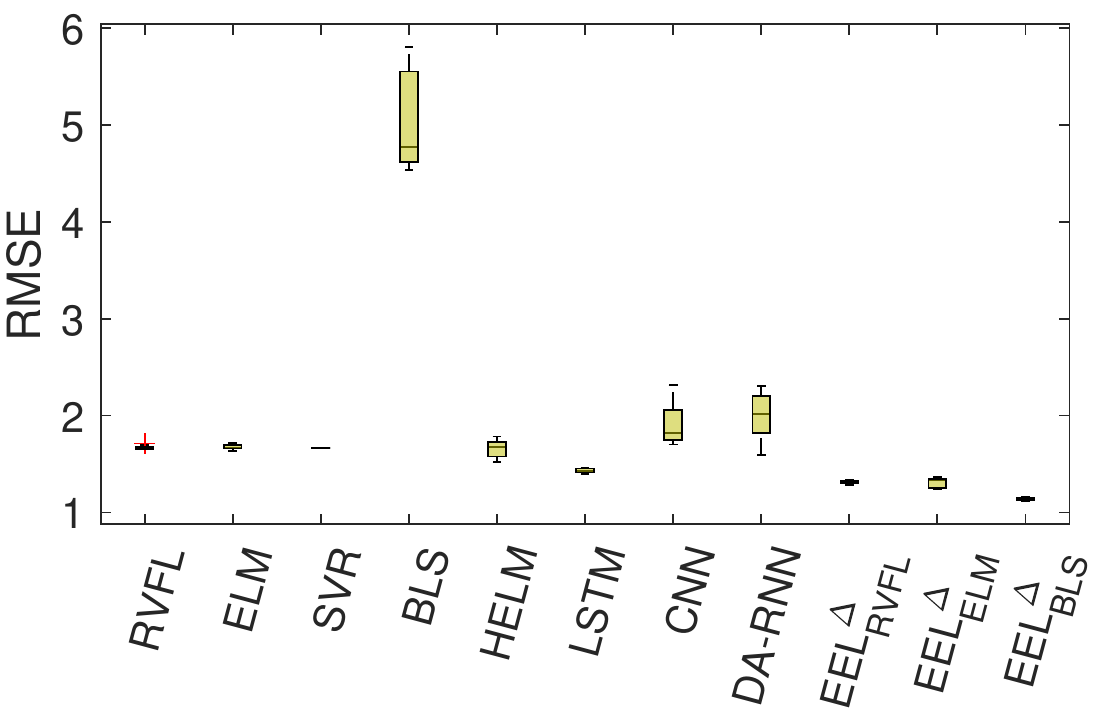}}\label{allcom_A}}\hspace{-0.1cm}
		\subfigure[Dataset B]
		{\centering\scalebox{0.39}[0.39]
			{\includegraphics{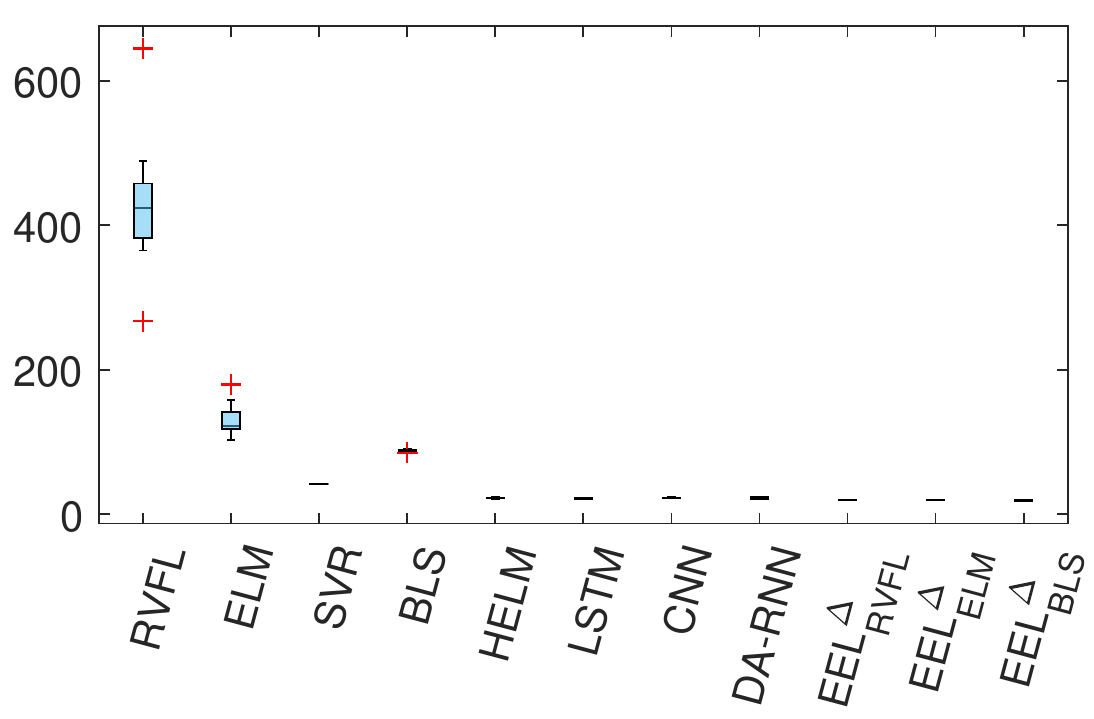}}\label{allcom_B}}
		\caption{Box plots of RVFL, ELM, SVR, BLS, HELM, LSTM, CNN, DA-RNN \cite{qin2017dual}, $\text{EEL}_\text{RVFL}^\Delta$, $\text{EEL}_\text{ELM}^\Delta$, $\text{EEL}_\text{BLS}^\Delta$.} \label{allcom}
	\end{figure}
	\subsubsection{Comparison with State-of-The-Art Models}
	
	Tab.~\ref{all_com} presents the average RMSE of RVFL, ELM, SVR, BLS, HELM, LSTM, CNN, DA-RNN \cite{qin2017dual}, $\text{EEL}_\text{RVFL}^\Delta$, $\text{EEL}_\text{ELM}^\Delta$ and $\text{EEL}_\text{BLS}^\Delta$ over 10 runs on each dataset and the labelled bold value represents better performance than the comparison models. The results of $\text{EEL}_\text{RVFL}^\Delta$, $\text{EEL}_\text{ELM}^\Delta$ and $\text{EEL}_\text{BLS}^\Delta$ are obtained from the best accuracy over each base learner in Tab.~\ref{comparison_EEL}. From the mean RMSE, we can see EEL (i.e., $\text{EEL}_\text{RVFL}^\Delta$, $\text{EEL}_\text{ELM}^\Delta$ and $\text{EEL}_\text{BLS}^\Delta$) has better prediction performance than comparison models on both datasets and $\text{EEL}_\text{BLS}^\Delta$ leads to the highest prediction accuracy. Statistical test is applied to show the difference of the results between the proposed EEL and the comparison models. The labelled bold and italic value means that it outperforms other models in the table significantly. It is obvious to notice that $\text{EEL}_\text{BLS}^\Delta$ outperforms RVFL, ELM, SVR, BLS, HELM, LSTM, CNN, and DA-RNN on both datasets. From the statistical result, we can observe that $\text{EEL}_\text{RVFL}^\Delta$, $\text{EEL}_\text{ELM}^\Delta$ and $\text{EEL}_\text{BLS}^\Delta$ have same prediction performance on the significance level for Dataset B and $\text{EEL}_\text{BLS}^\Delta$ outperforms $\text{EEL}_\text{RVFL}^\Delta$ and $\text{EEL}_\text{ELM}^\Delta$ for Dataset A. Fig.~\ref{allcom_A} and Fig.~\ref{allcom_B} show the box plots of RMSE over 10 runs obtained by RVFL, ELM, SVR, BLS, HELM, LSTM, CNN,  DA-RNN, $\text{EEL}_\text{RVFL}^\Delta$, and $\text{EEL}_\text{ELM}^\Delta$, $\text{EEL}_\text{BLS}^\Delta$. It tells that the proposed EEL over multiple PFs (i.e., $\text{EEL}^\Delta$) consistently demonstrates robust performance as evidenced by the smaller variance.
	
	\section{Conclusions and Future Work}\label{Conclusions}
	
	In this study, we propose a novel EEL framework based on SEL to simultaneously optimize the MTS prediction pipeline composed of CS, FE and PM, where several optimization tasks are involved, i.e., selection of channels, FE methods, TW, the suitable predictor and its parameters for configuration. RVFL, ELM and BLS are chosen as the base learners given their super-fast computational ability and promising performance. EEL employs MOEA/D subjected to two conflicting objectives, i.e., accuracy and diversity, to search for the PF consisted of a set of optimal solutions. By setting different population sizes in MOEA/D, a set of PFs are produced. SEL methods such as SBS+LS, and SFS+LS are designed for the ensemble modelling. EEL with multiple evolved PFs leads to better performance than the single PF. Compared with several state-of-the-art models, the superiority of the proposed novel EEL is further verified on two real-world MTS prediction. The computation time for the proposed EEL is an issue to be considered further and accordingly, our future work includes the investigation of parallel implementations of EEL. We are also interested to extend our work to multi-step prediction by using multi-task learning scenario \cite{chandra2017co,ZHENG202016}. 
	
	\bibliographystyle{IEEEtran}
	\bibliography{hui}

\end{document}